%% file: main.tex
\documentclass{article}

\usepackage[final]{neurips_2023}

\usepackage[utf8]{inputenc} % allow utf-8 input
\usepackage[T1]{fontenc}    % use 8-bit T1 fonts
\usepackage{hyperref}       % hyperlinks
\usepackage{url}            % simple URL typesetting
\usepackage{booktabs}       % professional-quality tables
\usepackage{amsfonts}       % blackboard math symbols
\usepackage{nicefrac}       % compact symbols for 1/2, etc.
\usepackage{microtype}      % microtypography
\usepackage{graphicx}
\usepackage{amsmath}
\usepackage{algorithm}
\usepackage{algpseudocode}
\usepackage{xspace}
\usepackage{subcaption}
\usepackage{caption}
\usepackage{tikz}
\algrenewcommand\algorithmicrequire{\textbf{Input:}}
\algrenewcommand\algorithmicensure{\textbf{Output:}}
\usepackage[frozencache,cachedir=.]{minted}
\usepackage{listings}
\usepackage{xcolor}
\usepackage{bold-extra}

\definecolor{darkblue}{rgb}{0.0,0.0,0.65}
\definecolor{darkred}{rgb}{0.68,0.05,0.0}
\definecolor{darkgreen}{rgb}{0.0,0.29,0.29}
\definecolor{darkpurple}{rgb}{0.47,0.09,0.29}
\definecolor{lightgray}{rgb}{0.95, 0.95, 0.95}

\newcommand{\eg}{\textit{e.g.}}

\newcommand*{\img}[1]{%
    \raisebox{-.1\baselineskip}{%
        \includegraphics[
        height=\baselineskip,
        width=\baselineskip,
        keepaspectratio,
        ]{#1}%
    }%
}
\definecolor{keywordcolor}{RGB}{0,0,128}
\definecolor{commentcolor}{RGB}{0,128,0}
\definecolor{stringcolor}{RGB}{163,21,21}
\usepackage{hyperref}
\hypersetup{
   colorlinks = true,
   citecolor  = darkblue,
   linkcolor  = darkred,
   filecolor  = darkblue,
   urlcolor   = darkblue,
 }

\title{Relational Deep Learning: Graph Representation Learning on Relational Databases}

\newcommand{\AreaName}{Relational Deep Learning\xspace}
\newcommand{\BenchmarkName}{\textsc{RelBench}\xspace}

\newcommand{\Amazon}{\texttt{rel-amazon}\xspace}
\newcommand{\AmazonChurn}{\texttt{rel-amazon-churn}\xspace}
\newcommand{\AmazonLTV}{\texttt{rel-amazon-ltv}\xspace}
\newcommand{\StackEx}{\texttt{rel-stackex}\xspace}
\newcommand{\StackExVotes}{\texttt{rel-stackex-votes}\xspace}
\newcommand{\StackExEngage}{\texttt{rel-stackex-engage}\xspace}

\newcommand{\tbl}[1]{\textsc{#1}\xspace} %relational table
\newcommand{\xhdr}[1]{\paragraph{#1.}\xspace} %column in a relational table

\newcommand{\jure}[1]{\textcolor{red}{\textbf{JL:}{ #1}}}

\usepackage{authblk}

\renewcommand\and{}

\author[2,*]{\textbf{Matthias Fey}}
\author[2,*]{\textbf{Weihua Hu}}
\author[1,*]{\textbf{Kexin Huang}}
\author[2,3,*]{\textbf{Jan Eric Lenssen}}
\author[1,*]{\textbf{Rishabh Ranjan}}
\author[1,*]{\\\textbf{Joshua Robinson}}
\author[2,4]{\textbf{Rex Ying}}
\author[2,5]{\textbf{Jiaxuan You}}
\author[1,2]{, \textbf{Jure Leskovec}}

\affil[*]{Equal contribution. Listed in alphabetical order. \vspace{1em}}

\affil[1]{Stanford University} 

\affil[2]{Kumo.AI}
\affil[3]{Max Planck Institute for Informatics}
\affil[4]{Yale University}
\affil[5]{University of Illinois at Urbana-Champaign \vspace{1em}}

\affil[ ]{\BenchmarkName: \href{https://relbench.stanford.edu/}{https://relbench.stanford.edu}}

\begin{document}

\maketitle

\begin{abstract}
Much of the world's most valued data is stored in relational databases and data warehouses, where the data is organized into many tables connected by primary-foreign key relations. 
However, building machine learning models using this data is both challenging and time consuming. 
The core problem is that no machine learning method is capable of learning on multiple tables interconnected by primary-foreign key relations.
Current methods can only learn from a single table, so the data must first be manually joined and aggregated into a single training table, the process known as feature engineering. Feature engineering is slow, error prone and leads to suboptimal models.
Here we introduce an end-to-end deep representation learning approach to directly learn on data laid out across multiple tables. We name our approach {\em \AreaName (RDL)}.
The core idea is to view relational databases as a temporal, heterogeneous graph, with a node for each row in each table, and edges specified by primary-foreign key links. Message Passing Graph Neural Networks can then automatically learn across the graph to extract representations that leverage all input data, without any manual feature engineering. \AreaName leads to more accurate models that can be built much faster.
To facilitate research in this area, we develop {\em \BenchmarkName}, a set of benchmark datasets and an implementation of \AreaName. The data covers a wide spectrum, from discussions on Stack Exchange to book reviews on the Amazon Product Catalog.
Overall, we define a new research area that generalizes graph machine learning and broadens its applicability to a wide set of AI use cases.
\end{abstract}

\input{intro}

\input{problem-formulation}

\input{db-learning}

\input{benchmark}

\input{new-challenges}

\input{related}

\input{conclusion}

\newpage

\xhdr{Acknowledgments}  We thank Shirley Wu for useful discussions as we were selecting datasets to adopt. We gratefully acknowledge the support of DARPA under Nos. N660011924033 (MCS); NSF under Nos. OAC-1835598 (CINES), CCF-1918940 (Expeditions), Stanford Data Science Initiative, Wu Tsai Neurosciences Institute, Chan Zuckerberg Initiative, Amazon, Genentech, GSK, Hitachi, Juniper Networks, and KDDI.

\bibliography{bib}
\bibliographystyle{plainnat}

\end{document}

%% file: intro.tex
\section{Introduction}

The information age is driven by data stored in ever-growing relational databases and data warehouses 
that have come to underpin nearly all technology stacks. Data warehouses typically store information in multiple tables, with entities/rows in different tables connected using primary-foreign key relations, and managed using powerful query languages such as SQL \citep{codd1970relational,chamberlin1974sequel}. For this reason, data warehouses underpin many of today's large information systems, including e-commerce, social media, banking systems, healthcare, manufacturing, and open-source scientific knowledge repositories~\citep{johnson2016mimic,pubmed}.

Many predictive problems over relational data have significant implications for human decision making. A hospital wants to predict the risk of discharging a patient; an e-commerce company wishes to forecast future sales of each of their products; a telecommunications provider wants to predict which customers will churn; and a music streaming platform must decide which songs to recommend to a user. Behind each of these tasks is a rich relational schema of relational tables, and many machine learning models are built using this data~\citep{kaggle-survey}. 

\begin{figure}[t] % 'h' for here
  \centering
  \input{figs/f1_labels_alt}
  \caption{\textbf{\AreaName solves predictive tasks on relational data with end-to-end learnable models.} There are three main steps. (a) A relational database with multiple tables connected by primary-foreign keys is given. (b) A predictive task is specified and added to the database by introducing an additional training table. (c) Relational data is transformed into its {\em Relational Entity Graph}, and a Graph Neural Network is trained over the graph with the supervision provided by the training table. The predictive task can be node level (as in this illustration), link level (pairs of nodes), or higher-order.}
  \label{fig: high level outline (fig1)}
\end{figure}
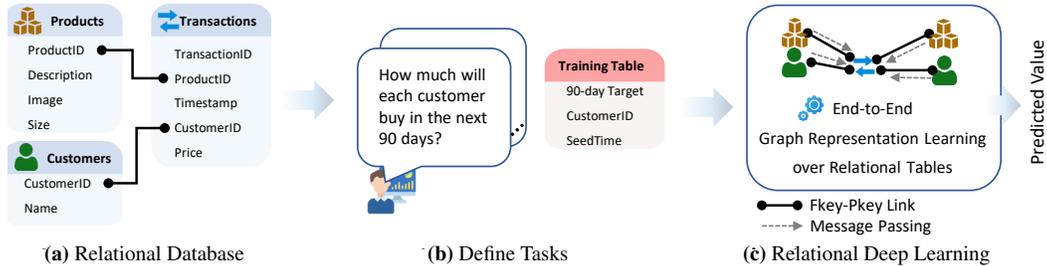

However, existing learning paradigms, notably tabular learning, cannot be directly applied to an interlinked set relational tables.
Instead, a manual feature engineering step is first taken, where a data scientist uses domain knowledge to manually join and aggregate tables to generate many features in a regular single table format. 
To illustrate this, consider a simple e-commerce schema (Fig. \ref{fig: high level outline (fig1)}) of three tables: \tbl{Customers}, \tbl{Transactions} and \tbl{Products}, where \tbl{Customers} and \tbl{Products} tables link into the \tbl{Transactions} table via primary-foreign keys, and the task is to predict if a customer is going to churn (\emph{i.e.}, make zero transactions in the next $k$ days). In this case, the data scientist would aggregate information from the \tbl{Transactions} table to make new features for the \tbl{Customers} table such as: ``number of purchases of a given customer in the last 30 days'', ``sum of purchase amounts of a given customer in the last 30 days'', ``number of purchases on a Sunday'', ``sum of purchase amounts on a Sunday'', ``number of purchases on a Monday'', and so on. The computed customer features are then stored in a single table, ready for tabular machine learning. Another challenge is the {\em temporal} nature of the churn predictive tasks. As new transactions appear, the customer's churn label and the customer's features may change from day to day, so features need to be recomputed for each day. Overall, the temporal nature of relational databases adds computational cost and further complexity, which often results in bugs and information leakage including the so-called ``time travel'' \citep{kapoor2023leakage}.

There are several issues with with the above approach: (1) it is a manual, slow and labor intensive process; (2) feature choices are essentially arbitrary and likely highly-suboptimal; 
%and the data scientists hopes that hand engineered features will be correlated with the label they aim to predict; 
(3) only a small fraction of the overall space of possible features can be manually explored; (4) by forcing data into a single table, information is aggregated into lower-granularity features, thus losing out on valuable fine-grain signal; (5) whenever the data distribution changes or drifts, current features become obsolete and new features have to be manually reinvented.

Many domains have been in a similar position, including pre-deep-learning computer vision, where hand-chosen convolutional filters  (\emph{e.g.}, Gabor) were used to extract features, followed by models such as SVMs or nearest neighbor search \citep{varma2005statistical}. Today, in contrast, deep neural networks skip the feature engineering and learn directly on the raw pixels, which results in large gains in model accuracy.
More broadly, the deep learning revolution has had a huge impact in many fields, including computer vision~\citep{he2016deep,russakovsky2015imagenet}, natural language processing~\citep{vaswani2017attention,devlin2018bert,brown2020language}, and speech~\citep{hannun2014deep,amodei2016deep}, and has led to super-human performance in many tasks. In all cases, the key was to move from manual feature engineering and handcrafted systems to fully data-driven, end-to-end representation learning systems.
For relational data, this transition has not yet occurred, as existing tabular deep learning approaches still heavily rely on manual feature engineering. 
Consequently, there remains a huge unexplored opportunity.

Here we introduce \emph{\AreaName (RDL)}, a blueprint for fulfilling the need for an end-to-end deep learning paradigm for relational tables (Fig. \ref{fig: high level outline (fig1)}).
Through end-to-end representation learning, \AreaName fully utilizes the rich predictive signals available in relational tables. The core of RDL is to represent relational tables as a temporal, heterogeneous  {\em Relational Entity Graph}, where each row defines a node, columns define node features, and primary-foreign key links define edges. Graph Neural Networks (GNNs)~\citep{gilmer2017mpgnn, hamilton2017inductive} can then be applied to build end-to-end data-driven predictive models.

Predictive tasks are specified on relational data by introducing a {\em training table} that holds supervision label information, (Fig. \ref{fig: high level outline (fig1)}b) but no input features. 
Training tables have two critically important characteristics. First, labels can be automatically computed from historical relational data, without any need for outside annotation; second, they may contain any number of foreign keys, permitting many task types including entity level (1 key, as in Fig. \ref{fig: high level outline (fig1)}b), link-level tasks such as recommendation (2 keys) and multi-entity tasks ($>$2 keys). Training tables permit many different types of prediction targets, including multi-class, multi-label, regression and more, ensuring high task generality.

\begin{figure}[t] % 'h' for here
  \centering
  \input{figs/f3_labels_alt_2}
  \caption{\textbf{Relational Deep Learning Pipeline.} \textbf{(a)} Given relational tables and a predictive task, a training table, containing supervised label information, is constructed and attached to the entity table(s). \textbf{(b)} Relational tables contain individual entities that are linked by foreign-primary key relations. \textbf{(c)} Relational data can be viewed as a single {\em Relational Entity graph}, which has a node for each entity, and edges given by primary-foreign key links. \textbf{(d)} Initial node features are extracted from each row in each table using modality-specific neural networks. Then a message passing graph neural network computes relation-aware node embeddings, a model head produces predictions for training table entities, and errors are backpropogated.
  }
  \label{fig: end to end pipeline}
\end{figure}
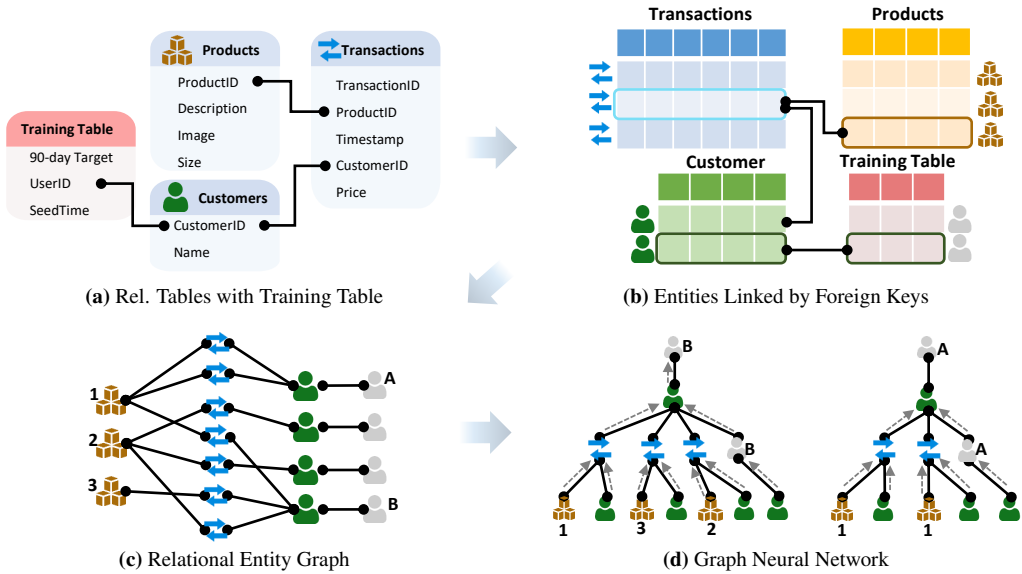

All in all, RDL model pipeline has four main steps (Fig.~\ref{fig: end to end pipeline}): Given a predictive machine learning task, (1)  A \emph{training table} containing supervision labels is constructed in a task-specific manner based on historic data in the relational database, (2) entity-level features are extracted and encoded from each row in each table to serve as node features, (3) node representations are learned through an inter-entity message-passing GNN that exchanges information between entities with primary-foreign key links, (4) a task-specific model head produces predictions for training data, and errors are backpropogated through the network. 

%% Jure commented this out. It feels like the issue of temporality
 Crucially, RDL models natively integrate temporality by only allowing entities to receive messages from other entities with earlier timestamps. This ensures that learned representation is automatically updated during GNN forward pass when new data is collected, and prevents information leakage and time travel bugs. Furthermore, this also stabilizes the generalization across time since models are trained to make predictions at multiple time snapshots by dynamically passing messages between entities at different time snapshots, whilst remaining grounded in a single relational database.

\xhdr{\BenchmarkName} To facilitate research into \AreaName, we introduce {\em \BenchmarkName}, a benchmarking and an evaluation Python package. Data in \BenchmarkName cover rich relational databases from many different domains. \BenchmarkName has the following key modules (1) \textbf{Data:} data loading, specifying a predictive task, and (temporal) data splitting, (2) \textbf{Model:} transforming data to a heterogeneous graph, building graph neural network predictive models,  (3) \textbf{Evaluation:} standardized evaluation protocol given a file of predictions. Importantly, data and evaluation modules are deep learning framework agnostic, enabling broad compatibility. To facilitate research, we provide our initial model implementation based on PyTorch Frame~\citep{Hu_PyTorch_Frame_A_2023} for encoding table rows into input node embeddings, which is then processed by GNN models in PyTorch Geometric~\citep{fey2019fast} to update the embeddings via message passing over the relational entity graph.

For the initial release, \BenchmarkName contains two databases, each with two predictive tasks. The first database is from Stack Exchange, the question-answering website, and includes 7 tables such as posts, users, and votes. The predictive tasks are (1) to predict if a user is going to make a new contribution (post, answer etc.), and (2) to predict the popularity of a new question. The second database is a subset of the Amazon Product Catalog focusing on books. There are three tables: users, products, and reviews. The tasks are (1) to predict the lifetime value of a user, and (2) whether a user will stop using the site. 

  \iffalse
  \jure{can we change this example to be in line with our running example (H\&M schema).}
  \jure{Here we have 3 steps, in the introduction we say there are 4 steps.}
  \jure{it is unclear what those icons are and what do they refer to -- 3 cubes, thumbs up/down. Basically, use these icons also on the left where you define the schema.}
  \jure{I think traning table needs to appear as an extra table on the left. On the right training table is represented as a separate node labeled that then points to the right entity. Right now the problem is that training table appears on the bottom right and connects to a graph.\\
  Here is how to update the Fig.: left tower: remove the "Machine Learning" and "Prediction", replace that with the schema (from the top) and add the greem training table to the schema. Right tower show how we go from schema to the relational graph (which includes nodes from the data tables as well as the training table). Let me know if this makes sense.}
  \jure{Sec. 2 shoudl refer ot the "left tower" and Sec. 3 should refer to the panels of the ``right tower'' of Fig. 1. Step by step. If this makes the Fig. too big/detailed, we should perhaps split Fig 1 into two separate large Fig.s. One Fig. explains the graph creation. The other Fig. explains the GNN workflow.}
  \fi

Our objective is to establish deep learning on relational data as a new subfield of machine learning. We hope that this will be a fruitful research direction, with many opportunities for impactful ideas that make much better use of the rich predictive signal in relational data. This paper lays the ground for future work by making the following main sections:
\begin{itemize}
    \item \textbf{Blueprint.} {\em \AreaName}, an end-to-end learnable approach that ultilizes the predictive signals available in relational data, and supports temporal predictions.
    \item \textbf{Benchmarking Package} {\em \BenchmarkName}, an open-source Python package for benchmarking and evaluating GNNs on relational data. {\BenchmarkName} beta release introduces two relational databases, and specifies two prediction tasks for each. 
    \item \textbf{Research Opportunities.} Outlining a new research program for  \AreaName, including multi-task learning, new GNN architectures, multi-hop learning, and more.
\end{itemize}

\xhdr{Organization} Section \ref{sec: problem scope} provides background on relational tables and predictive task specification. 
Section \ref{sec:formal_rl_problem} introduces our central methodological contribution, a graph neural network approach to solving predictive tasks on relational data.
Section \ref{sec: benchmark} introduces {\em \BenchmarkName}, a new benchmark for relational tables, and standardized evaluation protocols. 
Section \ref{sec: new challenges} outlines a landscape of new research opportunities for graph machine learning on relational data. Finally, Section \ref{sec: related work} concludes by contextualizing our new framework within the tabular and graph machine learning literature.

%% file: figs/f1_labels_alt.tex
\begin{tikzpicture}

\node (ex1) at (5,0)
{\includegraphics[trim={0.4cm 0.6cm 0cm 0cm}, clip, scale=0.62]{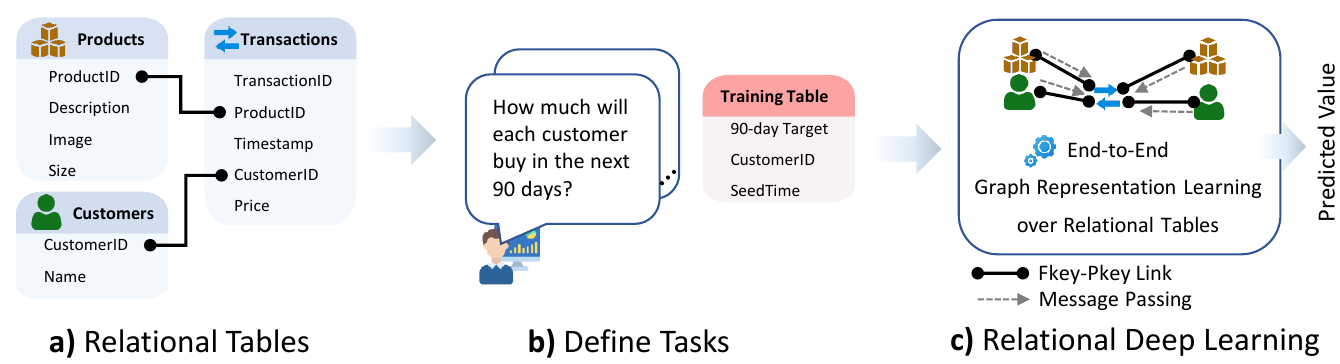}};
\node[scale=0.8] at (-0.1,-1.8) { \textbf{(a)} Relational Database};

\node[scale=0.8] at (4.6,-1.8) { \textbf{(b)} Define Tasks};

\node[scale=0.8] at (9.5,-1.81) { \textbf{(c)} Relational Deep Learning};

\end{tikzpicture}

%% file: figs/f3_labels_alt_2.tex
\begin{tikzpicture}

\node (ex1) at (4.7,0)
{\includegraphics[trim={0cm 0cm 0cm 0cm}, clip, scale=0.67]{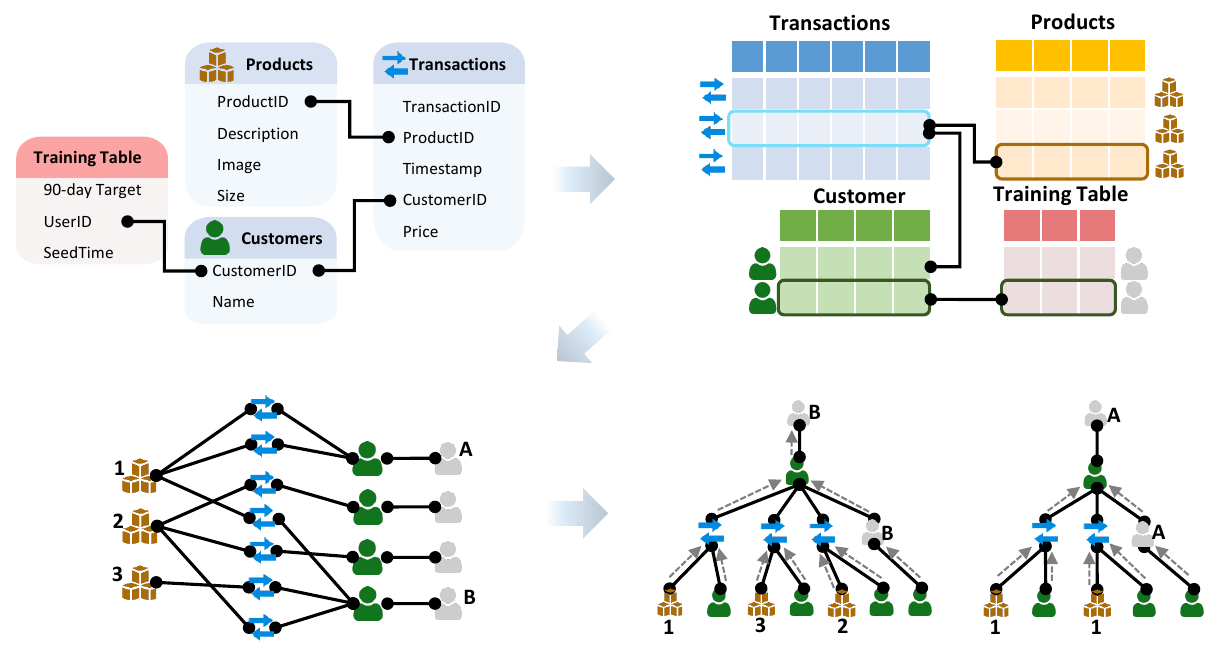}};
\node[scale=0.8] at (1,-0.3) { \textbf{(a)} Rel. Tables with Training Table};

\node[scale=0.8] at (8.2,-0.3) { \textbf{(b)} Entities Linked by Foreign Keys};

\node[scale=0.8] at (1,-3.8) { \textbf{(c)} Relational Entity Graph};

\node[scale=0.8] at (8.2,-3.8) { \textbf{(d)} Graph Neural Network};

\end{tikzpicture}

%% file: problem-formulation.tex
\section{Predictive Tasks on Relational Databases}\label{sec: problem scope}

This section outlines our problem scope: predictive tasks on relational tables. In the process, we define what we mean by relational tables, and how to specify predictive tasks on them. This section focuses exclusively on the structure of data and tasks, laying the groundwork for Section \ref{sec:formal_rl_problem}, which presents our GNN-based modelling approach.

\subsection{Relational Data}
\label{sec:rel_dbs}

\xhdr{A Brief History}
Relational tables and relational databases emerged in the 1970s as a means to standardize data retrieval and management \citep{codd1970relational}. As society digitized, relational databases came to fulfill a foundational purpose, and today are estimated to comprise 72\% of the world's data~\citep{db-engines}. Whilst there is no single agreed upon definition of a relational database, three essential characteristics are shared in all cases (\emph{cf.} Figure \ref{fig: high level outline (fig1)}a):
\begin{enumerate}
    \item Data is stored in multiple tables.
    \item Each row in each table contains an \emph{entity}, which possesses a unique primary key ID, along with multiple attributes stored as columns of the table.
    \item One entity may refer to another entity using a foreign key---the primary key of another entity.
\end{enumerate}

As well as a standardized storage system, relational databases typically come equipped with a powerful set of \emph{relational operations}, which are used to manipulate and access data. \cite{codd1970relational} introduced 8 relational operations, including set operations such as taking the union of two tables, and other operations such as \emph{joining} two tables based on their common attributes. Popular query languages such as SQL \citep{chamberlin1974sequel} provide commercial-grade implementations of a wide variety of relational operations.
Next we formally define relational data, as suits our purposes.

\xhdr{Definition of Relational Databases} A relational database $(\mathcal T, \mathcal L)$ is comprised of a collection of tables $\mathcal T = \{T_1 , \ldots , T_n\}$, and links between tables $\mathcal L \subseteq \mathcal T \times \mathcal T$ (\emph{cf.} Figure \ref{fig: end to end pipeline}a). A link $L = (T_{\rm fkey}$, $T_{\rm pkey})$ between tables exists if a foreign key column in $T_{\rm fkey}$ points to a primary key column of $T_{\rm pkey}$.
Each table is a set $T = \{v_1, ..., v_{n_T}\}$, whose elements $v_i \in T$ are called rows, or entities (\emph{cf.} Figure \ref{fig: end to end pipeline}b). Each entity $v \in T$, has four constituent parts $v = (p_v, \mathcal K_v, x_v, t_v)$: 
\begin{enumerate}
    \item \textbf{Primary key} $p_v$, that uniquely identifies the entity $v$. 
    \item \textbf{Foreign keys} $\mathcal K_v\subseteq \{p_{v'}:  v' \in T'\text{ and } (T, T') \in \mathcal L \}$, defining links between element $v\in T$ to elements $v' \in T'$, where $p_{v'}$ is the primary key of an entity $v'$ in table $T'$. 
    \item \textbf{Attributes} $x_v$, holding the informational content of the entity. % \matthias{can we define $d_T$?}
    \item \textbf{Timestamp} An optional timestamp $t_v$, indicating the time an event occurred. 
\end{enumerate}

For example, the \tbl{Transactions} table in Figure \ref{fig: end to end pipeline}a has the primary key (\tbl{TransactionID}), two foreign keys (\tbl{ProductID} and \tbl{CustomerID}), one attribute (\tbl{Price}), and timestamp column (\tbl{Timestamp}). Similarly, the \tbl{Products} table has the primary key (\tbl{ProductID}), no foreign keys, attributes (\tbl{Description}, \tbl{Image} and \tbl{Size}), and no timestamp. The connection between foreign keys and primary keys is illustrated by black connecting lines in Figure \ref{fig: end to end pipeline}.

In general, the attributes in table $T$ contain $d_T$ values: $x_v= (x_v^1, \ldots , x_v^{d_T})$, each belonging to a particular column. Critically, all entities in the same table have the same columns (values may be absent). Formally, this is described by membership  $x_v= (x_v^1, \ldots , x_v^{d_T}) \in \mathcal A_T^1 \times \ldots \times \mathcal A_T^{d_T}$, where $\mathcal A^i_T$ denotes the value space of $i$-th column of table $T$, and is shared between all entities $v \in T$.
%(such as numerical value, categorical value, image, or text spaces)
For example, the \tbl{Products} table from Fig.~\ref{fig: end to end pipeline}a contains three different attributes: the \emph{product description} (text type), the \emph{image} of the product (image type), and the \emph{size} of the product (numerical type). Each of these types has their own encoders as discussed in Sec.~\ref{sec:encoders}.

\xhdr{Fact and Dimension Tables} Tables are categorized into two types, \emph{fact} or \emph{dimension},  with complementary roles \citep{ullman-book}. Dimension tables provide contextual information, such as biographical information, macro statistics (such as number of beds in a hospital), or immutable properties, such as the size of a product (as in the \tbl{Products} table in Figure \ref{fig: end to end pipeline}a). Dimension tables tend to have relatively few rows, as it is limited to one per real-world object. Fact tables record interactions between other entities, such as all patient admissions to hospital, or all customer transactions (as in the \tbl{Transactions} table in Figure \ref{fig: end to end pipeline}a). Since entities can interact repeatedly, fact tables often contain the majority of rows in a relational database. Typically, features in dimension tables are static over their whole lifetime, while fact tables usually contain temporal information with a dedicated time column that denotes the time of appearance.

\xhdr{Temporality as a First-Class Citizen}

Relational data evolves over time as events occur and are recorded. 
This is captured by the (optional) timestamp $t_v$ attached to each entity $v$. For example, each transaction in \tbl{Transactions} has a time stamp.
Furthermore, many \emph{tasks} of interest involve forecasting future events. For example, how much will a customer spend in next $k$ days. It is therefore essential that time is conferred a special status unlike other attributes.  Our formulation, introduced in Section \ref{sec:formal_rl_problem}, achieves this through a temporal message passing scheme (similar to \cite{rossi2020temporal}), that only permits nodes to receive messages from neighbors with earlier timestamps. This ensures that models do not leak information from the future during training, avoiding shortcut decision rules that achieve high training accuracy but fail at test time \citep{geirhos2020shortcut}. It also means that model-extracted features are automatically updated as new relational data is added.

\subsection{From Task to Training Table}
\label{sec:predictive_labels}

There are many practically interesting machine learning tasks defined over relational databases, such as predicting the response of a patient to treatment, or the future sales of a product.
These tasks involve predicting the future state of the entities of interest.
Given a task we wish to solve, how can we create \emph{ground truth labels} to supervise machine learning models training? 

Our key insight is that we can generate training labels using historical data.
For instance, at time $t$, ground truth labels for predicting ``how much each customer will buy in the next $90$ days?'' are computed by summing up each customer's spending within the interval $t$ and $t+90$ days. Importantly, as long as $t+90$ is less than the most recent timestamp in the database, then these ground truth labels can be computed \emph{purely from historical data} without any need for external annotation. Further, by choosing different time points $t$ across the database time horizon, it is possible to naturally compute many ground truth training labels for each entity.

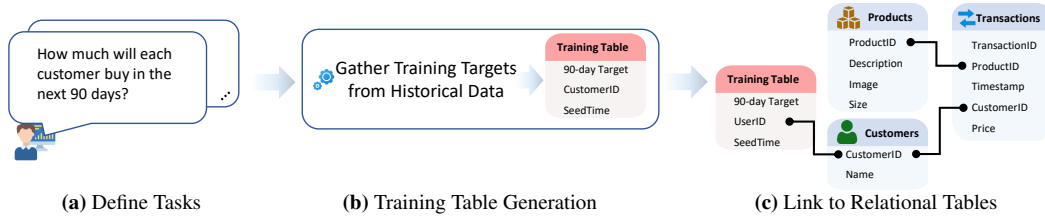
\begin{figure}[t] % 'h' for here
  \centering
    \input{figs/f2_labels}\vspace{-0.4cm}
  \caption{\textbf{Predictive Task Definition.} A task over relational data is defined by attaching an additional \emph{training table} to the existing linked tables. A training table entity specifies \textbf{(a)} ground truth label computed from historical information \textbf{(b)} the entity ID(s) the labels correspond to, and \textbf{(c)} a timestamp that controls what data the model can use to predict this label.}
  \label{fig:training_table}
\end{figure}

To hold the labels for a new predictive task, we introduce a new table known as a \emph{training table} $T_\text{train}$ (Fig.~\ref{fig:training_table}).
 Each entity $v=(\mathcal K_v,t_v,y_v)$ in the training table $T_\text{train}$ has three components: (1) A (set of) foreign keys $\mathcal K_v$ indicating the entities the training example is associated to, (2) a timestamp $t_v$, and (3) the ground truth label itself $y_v$.  In contrast to tabular learning settings, the training table does \emph{not} contain input data $x_v$. The training table is linked to the main relational database $(\mathcal T, \mathcal L)$ by updating: (1)  the set of tables to $\mathcal T  \cup \{{T_\text{train}}\}$, and (2) the links between tables to $ \mathcal L \cup \mathcal{L}_{T_\text{train}}$, where $\mathcal{L}_{T_\text{train}}$ specifies tables that training table keys $\mathcal K_v$ point to.

As discussed in Sec. \ref{sec:rel_dbs}, careful handling of what data the model sees during training is crucial in order to ensure temporal leakage does not happen. This is achieved using the training timestamp. When the model is trained to output target $y_v$ for entity $v$ with timestamp $t_v$, temporal consistency is ensured by only permitting the model to receive input information from entities $u$ with timestamp $t_u \leq t_v$ (see Sec. \ref{sec:temporal_sampling} for details on training sampling).

Thus, the purpose of the training table is twofold: to specify training \emph{inputs} and \emph{outputs} of the machine learning model. First, it provides supervision on the model \emph{output} by specifying the the entities and their target training labels. Second, in the case of temporal tasks, the training table specifies the model \emph{input} by specifying the timestamp at which each historical training label is generated.

This training table formulation can model a wide range of predictive tasks on relational databases:
\begin{itemize}
    \item {\bf Node-level prediction tasks} (\eg, multi-class classification, multi-label classification, regression): The training table has three columns \tbl{(EntityID, Label, Time)}, indicating the foreign key, target label, and timestamp columns, respectively.
    \item {\bf Link prediction tasks}: The training table has columns \tbl{(SourceEntityID, TargetEntityID, Label, Time)}, indicating the foreign key columns for the source/target nodes, and the target label, and timestamp, respectively.
    \item \textbf{Temporal and static prediction tasks}: Temporal tasks make predictions about the future (and require a seed time), while non-temporal tasks impute missing values (\tbl{Time} is dropped).
\end{itemize}

\xhdr{Training Table Generation} In practice, training tables can be computed using time-conditioned SQL queries from historic data in the database. Given a query that describes the prediction targets for all prediction entities, e.g. the sum of sells grouped by products, from time $t$ to time $t+\delta$ in the future, we can move $t$ back in time in fixed intervals to gather historical training, validation and test targets for all entities (\emph{cf.} Fig.~\ref{fig:training_table}b). We store $t$ as timestamp for the targets gathered in each step.

%% file: figs/f2_labels.tex
\begin{tikzpicture}

\node (ex1) at (5,0)
{\includegraphics[trim={0.2cm 0.8cm 0cm 0cm}, clip, scale=0.52]{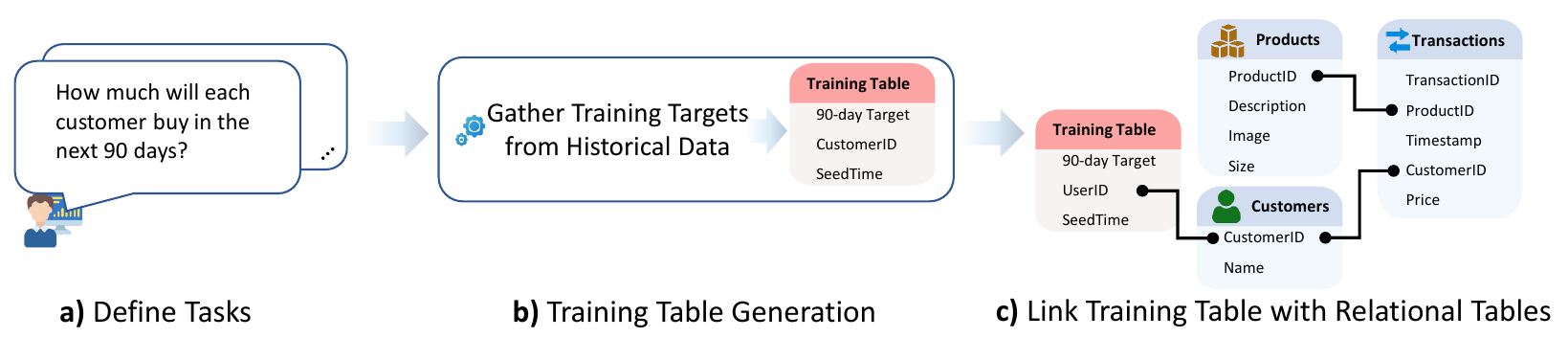}};
\node[scale=0.8] at (-0.5,-1.5) { \textbf{(a)} Define Tasks};

\node[scale=0.8] at (4,-1.51) { \textbf{(b)} Training Table Generation};

\node[scale=0.8] at (9.4,-1.5) { \textbf{(c)} Link to Relational Tables};

\end{tikzpicture}

%% file: db-learning.tex
\section{Predictive Tasks as Graph Representation Learning Problems}\label{sec:formal_rl_problem}

\begin{figure}[t] % 'h' for here
  \centering
   \centering
     \begin{subfigure}[b]{0.28\textwidth}
         \centering
         \includegraphics[width=\textwidth]{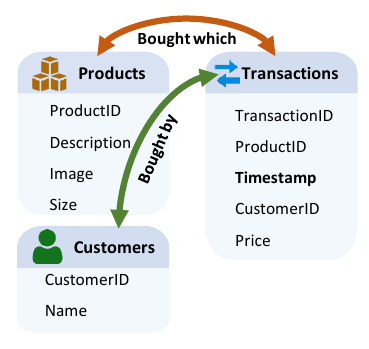}
         \vspace{-0.6cm}
         \caption{Schema Graph}
         \label{fig:schema_graph}
     \end{subfigure}
     \hfill
     \begin{subfigure}[b]{0.27\textwidth}
         \centering
         \includegraphics[width=\textwidth]{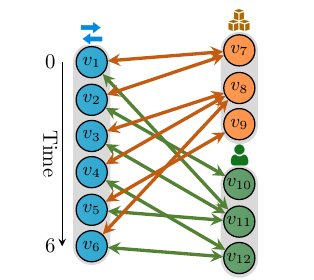}
         \caption{Relational Entity Graph}
         \label{fig:instrinsic_data_graph}
     \end{subfigure}
     \hfill
     \begin{subfigure}[b]{0.43\textwidth}
         \centering
         %\vspace{0.1cm}
         \includegraphics[width=\textwidth]{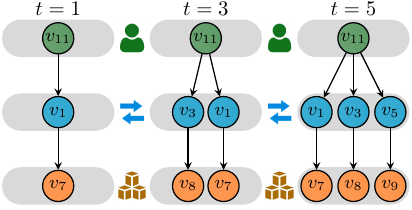}
         %\vspace{0.001cm}
         \caption{Computation Graphs for different time $t$}
         \label{fig:example_graph}
     \end{subfigure}

  \caption{\textbf{Three different kinds of graphs.} \textbf{(a)} The schema graph arises from the given relational tables. Each node denotes a table, and an edge between tables indicates that primary keys in one are foreign keys in the other. \textbf{(b)} The entity graph has one node for each entity in each table, and edges given by primary-foreign key links. The entity graph is heterogeneous with node and edge types defined by the schema graph. The nodes have a timestamp (illustrated by arrow-of-time), originating from the timestamp column of the table.
  \textbf{(c)} Using a temporal sampling strategy and a task description in form of training table containing different time $s$, we obtain \emph{time-consistent computation graphs} as training examples that naturally respect temporal order and map well to parallel compute.
  }
  \label{fig:three_graphs}
\end{figure}

Here, we formulate a generic machine learning architecture based on Graph Neural Networks, which solves predictive tasks on relational databases.
The following section will first introduce three important graph concepts, which are outlined in Fig.~\ref{fig:three_graphs}: (a) The \emph{schema graph} (\emph{cf.}~Sec.~\ref{sec:schema_graph}), table-level graph, where one table corresponds to one node. (b) The \emph{relational entity graph} (\emph{cf.} Sec.~\ref{sec:formal_rl_graph}), an entity-level graph, with a node for each entity in each table, and edges are defined via foreign-primary key connections between entities. (c) The \emph{time-consistent computation graph} (\emph{cf.}~Sec.~\ref{sec:temporal_sampling}), which acts as an explicit training example for graph neural networks.
We describe generic procedures to map between graph types, and finally introduce our GNN blueprint for end-to-end learning on relational databases (\emph{cf.}~Sec.~\ref{sec:task_specific_gnn}).

\subsection{Schema Graph}
\label{sec:schema_graph}

The first graph in our blueprint is the \emph{schema graph} (\emph{cf.} Fig. \ref{fig:three_graphs}a), which describes the table-level structure of data.
Given a relational database $(\mathcal{T}, \mathcal{L})$ as defined in Sec.~\ref{sec: problem scope}, we let \mbox{$\mathcal L^{-1} = \{(T_{\rm pkey}, T_{\rm fkey}) \mid (T_{\rm fkey}, T_{\rm pkey}) \in \mathcal{L}\}$} denote its inverse set of links.
Then, the \emph{schema graph} is the graph $(\mathcal{T}, \mathcal{R})$ that arises from the relational database, with node set $\mathcal{T}$ and edge set $\mathcal{R} = \mathcal L \cup \mathcal L^{-1}$.
Inverse links ensure that all tables are reachable within the schema graph.
The schema graph nodes serve as type definitions for the heterogeneous relational entity graph, which we define next.

\subsection{Relational Entity Graph}
\label{sec:formal_rl_graph}

To formulate a graph suitable for processing with GNNs, we introduce the \emph{relational entity graph}, which has entity-level nodes and serves as the basis of the proposed relational learning framework.

Our relational entity graph is a \emph{heterogeneous graph}  $G=(\mathcal{V}, \mathcal{E}, \phi, \psi)$, with node set $\mathcal{V}$ and edge set $\mathcal{E} \subseteq \mathcal V \times \mathcal V$ and type mapping functions $\phi: \mathcal{V} \rightarrow \mathcal{T}$ and $\psi: \mathcal{E} \rightarrow \mathcal{R}$, where each node $v \in \mathcal{V}$ belongs to a \emph{node type} $\phi(v) \in \mathcal{T}$ and each edge $e \in \mathcal E$ belongs to an \emph{edge type} $\psi(e) \in \mathcal{R}$.
Specifically, the sets $\mathcal{T}$ and $\mathcal{R}$ from the schema graph define the node and edge types of our relational entity graph.

Given a schema graph $(\mathcal{T}, \mathcal{R})$ with tables $T = \{v_1, ..., v_{n_T}\}\in\mathcal{T}$ as defined in Sec.~\ref{sec: problem scope}, we define the node set in our relational entity graph as the union of all entries in all tables $\mathcal{V} = \bigcup_{T\in\mathcal{T}}T$.
Its edge set is then defined as
\begin{equation}
\mathcal{E} = \{(v_1, v_2) \in \mathcal{V} \times \mathcal{V} \mid  p_{v_2} \in \mathcal{K}_{v_1} \text{ or } p_{v_1} \in \mathcal{K}_{v_v}\}\textnormal{,}
\end{equation}
\emph{i.e.} the entity-level pairs that arise from the primary-foreign key relationships in the database. 
We equip the relational entity graph with the following key information:
\begin{itemize}
    \item \textbf{Type mapping functions} $\phi: \mathcal{V} \rightarrow \mathcal{T}$ and $\psi: \mathcal{E} \rightarrow \mathcal{R}$, mapping nodes and edges to respective elements of the schema graph, making the graph \emph{heterogeneous}. We set $\phi(v) = T$ for all $v\in T$ and $\psi(v_1,v_2) = (\phi(v_1), \phi(v_2)) \in \mathcal R$ if $(v_1,v_2) \in \mathcal E$.
    \item \textbf{Time mapping function} $\tau: \mathcal{V} \rightarrow \mathcal{D}$, mapping nodes to its timestamp: $\tau: v \mapsto t_v$ (as defined in Sec.~\ref{sec:rel_dbs}), introducing time as a central component and establishes the \emph{temporality} of the graph. The value $\tau(v)$ denotes the point in time in which the table row $v$ became available or $-\infty$ in case of non-temporal rows.
    \item \textbf{Embedding vectors} $\mathbf{h}_v \in \mathbb{R}^{d_{\phi(v)}}$ for each $v\in\mathcal{V}$, which contains an embedding vector for each node in the graph. Initial embeddings are obtained via multi-modal column encoders as described in Sec.~\ref{sec:encoders}. Final embeddings are computed via GNNs outlined in Section 3.4.
\end{itemize}

An example of a relational entity graph for a given schema graph is given in Fig.~\ref{fig:instrinsic_data_graph}.
The graph contains a node for each row in the database tables. Two nodes are connected if the foreign key entry in one table row links to the primary key entry of another table row. Node and edge types are defined by the schema graph. Nodes resulting from temporal tables carry the timestamp from the respective row, allowing temporal message passing, which is described next.

\subsection{Time-Consistent Computational Graphs}
\label{sec:temporal_sampling}

Given a relational entity graph and a training table  (\emph{cf.}~Sec.~\ref{sec:predictive_labels}), we need to be able to query the graph at specific points in time which then serve as explicit training examples used as input to the model.
In particular, we create a subgraph from the relational entity graph induced by the set of foreign keys $\mathcal{K}_v$ and its timestamp $t_v$ of a training example in the training table $T_\text{train}$.
This subgraph then acts as a local and \emph{time-consistent computation graph} to predict its ground-truth label $y_v$.

\begin{algorithm}[t]
\caption{Time-Consistent Computation Graph}\label{alg:temp_sampling}
\begin{algorithmic}
\Require Relational entity graph $G = (\mathcal{V}, \mathcal{E})$, number of hops $L$, seed node $v_0\in\mathcal{V}$, seed time $t \in \mathbb{R}$
\Require Neighborhood sizes $(m_1,...,m_L) \in \mathbb{N}^L$
\Ensure Computation graph $G_{\rm comp} = (\mathcal{V}_{\rm comp}, \mathcal{E}_{\rm comp})$
\State $\mathcal V_0 \gets \{v_0\}$\textnormal{,} \quad $\mathcal E_0 \gets \emptyset $
\For{$i \in \{1,...,L\}$}
\For{$v \in \mathcal V_{i-1}$}
    \State $\mathcal{E}_i \gets \textnormal{SELECT}_{m_i}(\{(w,v) \in \mathcal{E} \mid \tau(v)\leq t\})$ \Comment{Select a maximum of $m_i$ filtered edges}
    \State $\mathcal{V}_i \gets \{w \in \mathcal{V} \mid (w,v) \in \mathcal{E}_i\}$  \Comment{Gather nodes for the sampled edges}
\EndFor
\EndFor
\State $\mathcal V_{\rm comp} \gets \bigcup_{i=1}^L \mathcal V_i$\textnormal{,} \quad $\mathcal E_{\rm comp} \gets \bigcup_{i=1}^L \mathcal E_i$
\end{algorithmic}
\end{algorithm}

The computational graphs obtained via neighbor sampling~\citep{hamilton2017inductive} allow the scalability of our proposed approach to modern large-scale relational data  with billions of table rows, while ensuring the temporal constraints~\citep{wang2021temporalsampling}.
Specifically, given a number of hops $L$ to sample, a seed node $v \in \mathcal{V}$, and a timestamp $t$ induced by a training example, the computation graph is defined as $G_{\rm comp} = (\mathcal{V}_{\rm comp}, \mathcal{E}_{\rm comp})$ as the output of Alg.~\ref{alg:temp_sampling}. 
The algorithm traverses the graph starting from the seed node $v$ for $L$ iterations. In iteration $i$, it gathers a maximum of $m_i$ neighbors available up to timestamp $t$, using one of three selection strategies:
\begin{itemize}
    \item \textbf{Uniform temporal sampling} selects uniformly sampled random neighbors.
    \item \textbf{Ordered temporal sampling} takes the latest neighbors, ordered by time $\tau$.
    \item \textbf{Biased temporal sampling} selects random neighbors sampled from a multinomial probability distribution induced by $\tau$. For instance, sampling can be performed proportional to relative neighbor time or biased towards specific important historical moments. 
\end{itemize}

The temporal neighbor sampling is performed purely on the graph structure of the relational entity graph, without requiring initial embeddings $\mathbf{h}^{(0)}_v$. The bounded size of computation graph $G_{\rm comp}$ allows for efficient mini-batching on GPUs, independent of relational entity graph size.
In practice, we perform temporal neighbor sampling on-the-fly, which allows us to operate on a shared relational entity graph across all training examples, from which we can then restore local and historical snapshots very efficiently.
Examples of computation graphs are shown in Fig.~\ref{fig:example_graph}.

\subsection{Task-Specific Temporal Graph Neural Networks}
\label{sec:task_specific_gnn} 

Given a time-consistent computational graph and its future label to predict, we define a generic multi-stage deep learning architecture as follows:
\begin{enumerate}
\setlength\itemsep{0em}
    \item Table-level \textbf{column encoders} that encode table row data into initial node embeddings $\mathbf{h}^{(0)}_v$, as described in Sec.~\ref{sec:encoders}.
    \item A stack of $L$ \textbf{relational-temporal message passing layers} (\emph{cf.}~Sec.~\ref{sec:rl_mp}).
    \item A task-specific \textbf{model head}, mapping final node embeddings to a prediction (\emph{cf.}~Sec.~\ref{sec:model_heads}).
\end{enumerate}
The whole architecture, consisting of table-level encoders, message passing layers and task specific model heads can be trained end-to-end to obtain an optimal model for the given task.

\subsubsection{Relational-Temporal Message Passing}
\label{sec:rl_mp}
This section introduces a generic framework for heterogeneous message passing GNNs on relational entity graphs as defined in Sec.~\ref{sec:formal_rl_graph}.  
A message passing operator in the given relational framework needs to respect the heterogeneous nature as well as the temporal properties of the graph. This is ensured by filtering nodes based on types and time. Thus, we briefly introduce heterogeneous message passing before we turn to our temporal message passing.

\xhdr{Heterogeneous Message Passing} 

\emph{Message-Passing Graph Neural Networks (MP-GNNs)}~\citep{gilmer2017mpgnn, fey2019fast} are a generic computational framework to define deep learning architectures on graph-structered data. Given a heterogeneous graph $G=(\mathcal{V}, \mathcal{E}, \phi, \psi)$ with initial node embeddings $\{\mathbf{h}^{(0)}_v\}_{v\in\mathcal{V}}$, a single message passing iteration computes updated features $\{\mathbf{h}^{(i+1)}_v\}_{v\in\mathcal{V}}$ from features $\{\mathbf{h}^{(i)}_v\}_{v\in\mathcal{V}}$ given by the previous iteration.
One iteration takes the form: %computing embedding $\mathbf{h}^{i+1}_v$, by: 
\begin{equation}
\label{eq:mp}
   \mathbf{h}^{(i+1)}_v = f(\mathbf{h}^{(i)}_v, \{\hspace{-0.1cm}\{g(\mathbf{h}^{(i)}_w) \mid w \in \mathcal{N}(v)\}\hspace{-0.1cm}\})\textnormal{,}
\end{equation}
where $f$ and $g$ are arbitrary differentiable functions with optimizable parameters and $\{\hspace{-0.1cm}\{\cdot\}\hspace{-0.1cm}\}$ an permutation invariant set aggregator, such as mean, max, sum, or a combination. Heterogeneous message passing~\citep{schlichtkrull2018relational, hu2020hgt} is a \emph{nested} version of Eq.~\ref{eq:mp}, adding an aggregation over all incoming edge types to learn distinct message types:
\begin{equation}
\label{eq:hmp}
   \mathbf{h}^{(i+1)}_v = f_{\phi(v)}\Bigl(\mathbf{h}^{(i)}_v, \Bigl\{\hspace{-0.15cm}\Bigl\{ f_{R}( \{\hspace{-0.1cm}\{g_{R}(\mathbf{h}^{(i)}_w) \mid w \in \mathcal{N}_R(v)\}\hspace{-0.1cm}\}) \, \Big\vert \, \forall R=(T,\phi(v)) \in \mathcal{R}\Bigr\}\hspace{-0.15cm}\Bigr\}\Bigr)\textnormal{,}
\end{equation}
where $\mathcal{N}_R(v) =\{w\in \mathcal{V} \mid (w,v)\in \mathcal{E} \textnormal{ and } \psi(w, v) = R \}$ denotes the \emph{R-specific neighborhood} of node $v\in \mathcal{V}$.
This formulation supports a wide range of different graph neural network operators, which define the specific form of functions $f_{\phi(v)}$, $f_{R}$, $g_{R}$ and $\{\hspace{-0.1cm}\{\cdot\}\hspace{-0.1cm}\}$~\citep{fey2019fast}.
%We now introduce our temporal formulation.

\xhdr{Temporal Message Passing} Given a relational entity graph $G=(\mathcal{V}, \mathcal{E}, \mathcal{T}, \mathcal{R})$ with attached mapping functions $\psi, \phi, \tau$ and initial node embeddings $\{\mathbf{h}^{(0)}_v\}_{v\in\mathcal{V}}$ and an example specific \emph{seed time} $t\in \mathbb{R}$ (\emph{cf.} Sec.~\ref{sec:predictive_labels}) , we obtain a set of deep node embeddings $\{\mathbf{h}^{(L)}_v\}_{v\in\mathcal{V}}$ by $L$ consecutive applications of Eq.~\ref{eq:hmp}, where we additionally filter $R$-specific neighborhoods based on their timestamp, \emph{i.e.} replace $\mathcal{N}_R(v)$ with
\begin{equation}
     \mathcal{N}^{\leq t}_R(v) =\{w\in \mathcal{V} \mid (w,v)\in \mathcal{E} \textnormal{, } \psi(w, v) = R \textnormal{, and } \tau(w) \leq t\} \text{,}
\end{equation}
realized by the temporal sampling procedure presented in Sec.~\ref{sec:temporal_sampling}. The formulation naturally respects time by only aggregating messages from nodes that were available before the given seed time $s$. The given formulation is agnostic to specific implementations of message passing and supports a wide range of different operators.

\subsubsection{Prediction with Model Heads}
\label{sec:model_heads}

The model described so far is task-agnostic and simply propagates information through the relational entity graph to produce generic node embeddings. We obtain a task-specific model by combining our graph with a training table, leading to specific model heads and loss functions. We distinguish between (but are not limited to) two types of tasks: node-level prediction and link-level prediction.

\xhdr{Node-level Model Head}
Given a batch of $N$ node level training table examples $\{(\mathcal{K}, t, y)_i\}_{i=1}^N$ (\emph{cf.} Sec.~\ref{sec:predictive_labels}), where $\mathcal{K} = \{k\}$ contains the primary key of node $v~\in \mathcal V$ in the relational entity graph, $t\in \mathbb{R}$ is the seed time, and $y\in\mathbb{R}^d$ is the target value. Then, the node-level model head is a function that maps node-level embeddings $\mathbf{h}^{(L)}_{v}$ to a prediction $\hat{y}$, \emph{i.e.} 
\begin{equation}
    f: \mathbb{R}^{d_{v}} \rightarrow \mathbb{R}^d \textnormal{,} \quad \quad  f: \mathbf{h}^{(L)}_{v} \mapsto \hat{y} \textnormal{.}
\end{equation}
\xhdr{Link-level Model Head}
Similarly, we can define a link-level model head for training examples $\{(\mathcal K, t, y)_i\}_{i=1}^N$ with $\mathcal{K} = \{k_1, k_2\}$ containing primary keys of two different nodes $v_1, v_2 \in \mathcal{V}$ in the relational entity graph. A function maps node embeddings $\mathbf{h}^{(L)}_{v_1}$, $\mathbf{h}^{(L)}_{v_2}$ to a prediction, \emph{i.e.}
\begin{equation}
    f: \mathbb{R}^{d_{v_1}} \times \mathbb{R}^{d_{v_2}} \rightarrow \mathbb{R}^d \textnormal{,} \quad \quad f: (\mathbf{h}^{(L)}_{v_1}, \mathbf{h}^{(L)}_{v_2}) \mapsto \hat{y} \textnormal{.}
\end{equation}
A task-specific loss $L(\hat{y}, y)$ provides gradient signals to all trainable parameters. The presented approach can be generalized to $|\mathcal K|>2$ to specify subgraph-level tasks. In the first version, \BenchmarkName provides node-level tasks only.

\subsubsection{Multi-Modal Node Encoders}
\label{sec:encoders}
The final piece of the pipeline is to obtain the initial entity-level node embeddings $\mathbf{h}^{(0)}_v$ from the multi-modal input attributes $x_v= (x_v^1, \ldots , x_v^{d_T}) \in \mathcal{A}_T^1 \times \ldots \times \mathcal{A}_T^{d_T}$.
Due to the nature of tabular data, each column element $x_v^i$ lies in its own modality space $\mathcal{A}_T^{d_i}$, \eg, image, text, categorical, and numerical values.
Therefore, we use a modality-specific encoder to embed each attribute into embeddings.
For text and image modalities, we can naturally use pre-trained embedding models as the encoders~\citep{reimers2019sentence}.
After all the attributes are embedded, we apply state-of-the-art tabular deep learning models~\citep{huang2020tabtransformer,arik2021tabnet,gorishniy2021revisiting,gorishniy2022embeddings,chen2023trompt} to fuse all the attribute embeddings into a single entity-level node embedding.
In practice, we rely on PyTorch Frame~\citep{Hu_PyTorch_Frame_A_2023} that supports a variety of modality-specific encoders, such as pre-trained text embedding models, and as well as state-of-the-art deep learning models on tabular data.

\subsection{Discussion}

The neural network architecture presented in this blueprint is end-to-end trainable on relational databases. This approach supports a wide range of tasks, such as classification, regression or link prediction in a unified way, with labels computed and stored in a training table.  
It learns to solve tasks without requiring manual feature engineering, as typical in tabular learning. Instead, operations that are otherwise done manually, such as SQL \texttt{JOIN}+\texttt{AGGREGATE} 
 operations, are learned by the GNN. More than simply replacing SQL operations, the GNN message and aggregation steps \emph{exactly match} the functional form of SQL \texttt{JOIN}+\texttt{AGGREGATE} operations. In other words, the GNN is an exact \emph{neural version} of SQL \texttt{JOIN}+\texttt{AGGREGATE} operations. We believe this is another important reason why message passing-based architectures are a natural learned replacement for hand-engineered features on relational tables.

%% file: benchmark.tex
\section{\img{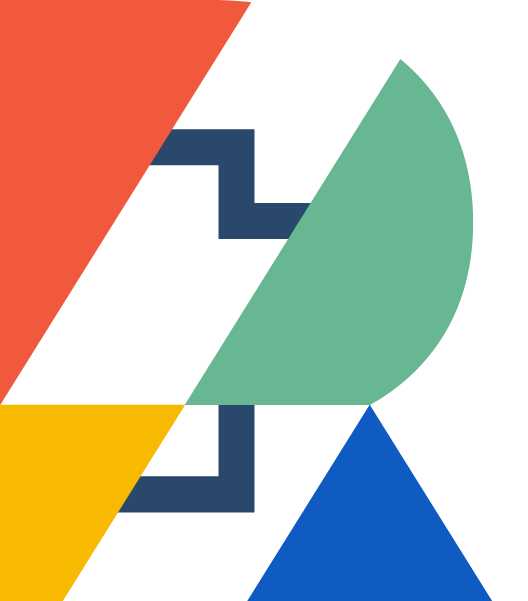}~\BenchmarkName: A Benchmark for \AreaName}\label{sec: benchmark}
\begin{figure}[h]
    \centering
    \includegraphics[width=\textwidth]{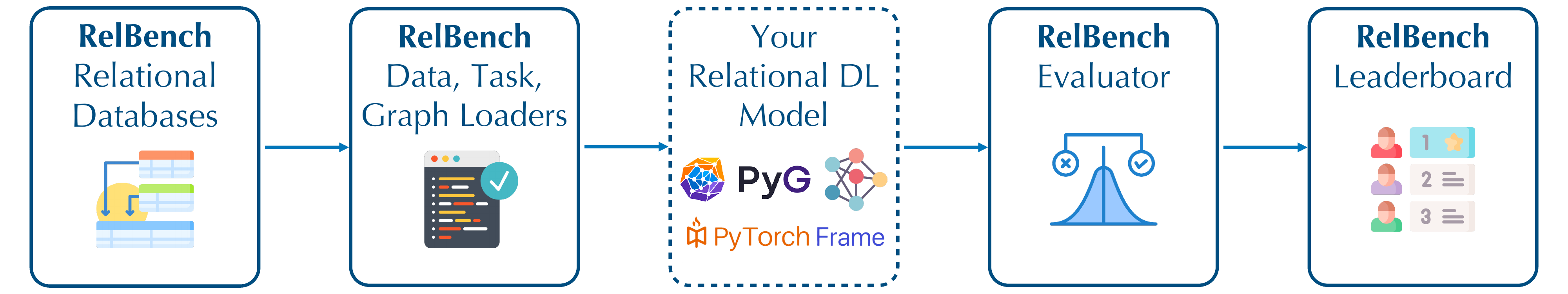}
    \caption{\textbf{Overview of \BenchmarkName.} \BenchmarkName enables training and evaluation of machine learning models on relational data. \BenchmarkName supports deep learning framework agnostic data loading, task specification, standardized data splitting, and transforming data into graph format. \BenchmarkName provides standardized evaluation metric computations, and a leaderboard for tracking progress.  We additionally provide example training scripts built using PyTorch Geometric and PyTorch Frame.}
    \label{fig:rtb-fig}
\end{figure}

We introduce \BenchmarkName, an open benchmark for \AreaName. The goal of \BenchmarkName is to facilitate scalable, robust, and reproducible machine learning research on relational tables. \BenchmarkName curates a diverse set of large-scale, challenging, and realistic benchmark databases and defines meaningful predictive tasks over these databases. In addition, \BenchmarkName develops a Python library for loading relational tables and tasks, constructing data graphs, and providing unified evaluation for predictive tasks. It also integrates seamlessly with existing Pytorch Geometric and PyTorch Frame functionalities. In its beta release\footnote{Website: \url{https://relbench.stanford.edu}}\footnote{Package: \url{https://github.com/snap-stanford/relbench}}, we announce the first two real-world relational databases, each with two curated predictive tasks. 

In the subsequent sections (Sec.~\ref{sec:amazon} and~\ref{sec:stack}), we describe in detail the two relational databases and the predictive tasks. For each database, we show its entity relational diagrams and important statistics. For each task, we define the task formulation, entity filtering, significance of the task, and also unified evaluation metric. Finally, we demonstrate the usage of the \BenchmarkName's package in Sec.~\ref{sec:package}.

\subsection{\BenchmarkName Package}\label{sec:package}

The \BenchmarkName package is designed to allow easy and standardized access to \AreaName for researchers to push the state-of-the-art of this emerging field. It provides Python APIs to (1) download and process relational databases and their predictive tasks; (2) load standardized data splits and generate relevant train/validation/test tables; (3) evaluate on machine learning predictions. It also provides a flexible ecosystems of supporting tools such as automatic conversion to PyTorch Geometric graphs and integration with Pytorch Frame to produce embeddings for diverse column types. We additionally provide end-to-end scripts for training using \BenchmarkName package with GNNs and XGBoost~\citep{chen2016xgboost}. We refer the readers to the code repository for a more detailed understanding of \BenchmarkName. Here we demonstrate the core functionality.

To load a relational database, simply do:

\vspace{-3mm}
\begin{minted}[frame=lines,
framesep=2mm,
baselinestretch=1,
fontsize=\footnotesize
]{python}
from relbench.datasets import get_dataset
dataset = get_dataset(name="rel-amazon")
\end{minted}
\vspace{-3mm}

It will load the relational tables and process it into a standardized format. Next, to load the predictive task and the relevant training tables, do:

\vspace{-3mm}
\begin{minted}[frame=lines,
framesep=2mm,
baselinestretch=1,
fontsize=\footnotesize
]{python}
task = dataset.get_task("rel-amazon-ltv")
task.train_table, task.val_table, task.test_table # training/validation/testing tables
\end{minted}
\vspace{-3mm}

It automatically constructs the training table for the relevant predictive task. Next, after the user trains the machine learning model, the user can use \BenchmarkName standardized evaluator:
\vspace{-3mm}
\begin{minted}[frame=lines,
framesep=2mm,
baselinestretch=1,
fontsize=\footnotesize
]{python}
task.evaluate(pred)
\end{minted}
\vspace{-3mm}

\subsection{Temporal Splitting}

Every dataset in \BenchmarkName has a validation timestamp $t_{\text{val}}$ and a test timestamp $t_{\text{test}}$. These are shared for all tasks in the dataset.
The test table for any task comprises of labels computed for the time window from $t_{\text{test}}$ to $t_{\text{test}} + \delta$, where the window size $\delta$ is specified for each task. Thus the model must make predictions using only information available up to time $t_{\text{test}}$. Accordingly, to prevent accidental temporal leakage at test time \BenchmarkName only provides database rows with timestamps up to $t_{\text{test}}$ for training and validation purposes. \BenchmarkName also provides default train and validation tables. The default validation table is constructed similar to the test table, but with the time window being $t_{\text{val}}$ to $t_{\text{val}} + \delta$. To construct the default training table, we first sample time stamps $t_i$ starting from $t_{\text{val}} - \delta$ and moving backwards with a stride of $\delta$. This allows us to benefit from the latest available training information. Then for each $t_i$, we apply an entity filter to select the entities of interest (e.g., active users). Finally for each pair of timestamp and entity, we compute the training label based on the task definition. Users can explore other ways of constructing the training or validation table, for example by sampling timestamps with shorter strides to get more labels, as long as information after $t_{\text{val}}$ is not used for training.

\subsection{\Amazon: Amazon product review e-commerce database}\label{sec:amazon}

\begin{figure}[h] % 'h' for here
  \centering
  \includegraphics[width=0.5\linewidth]{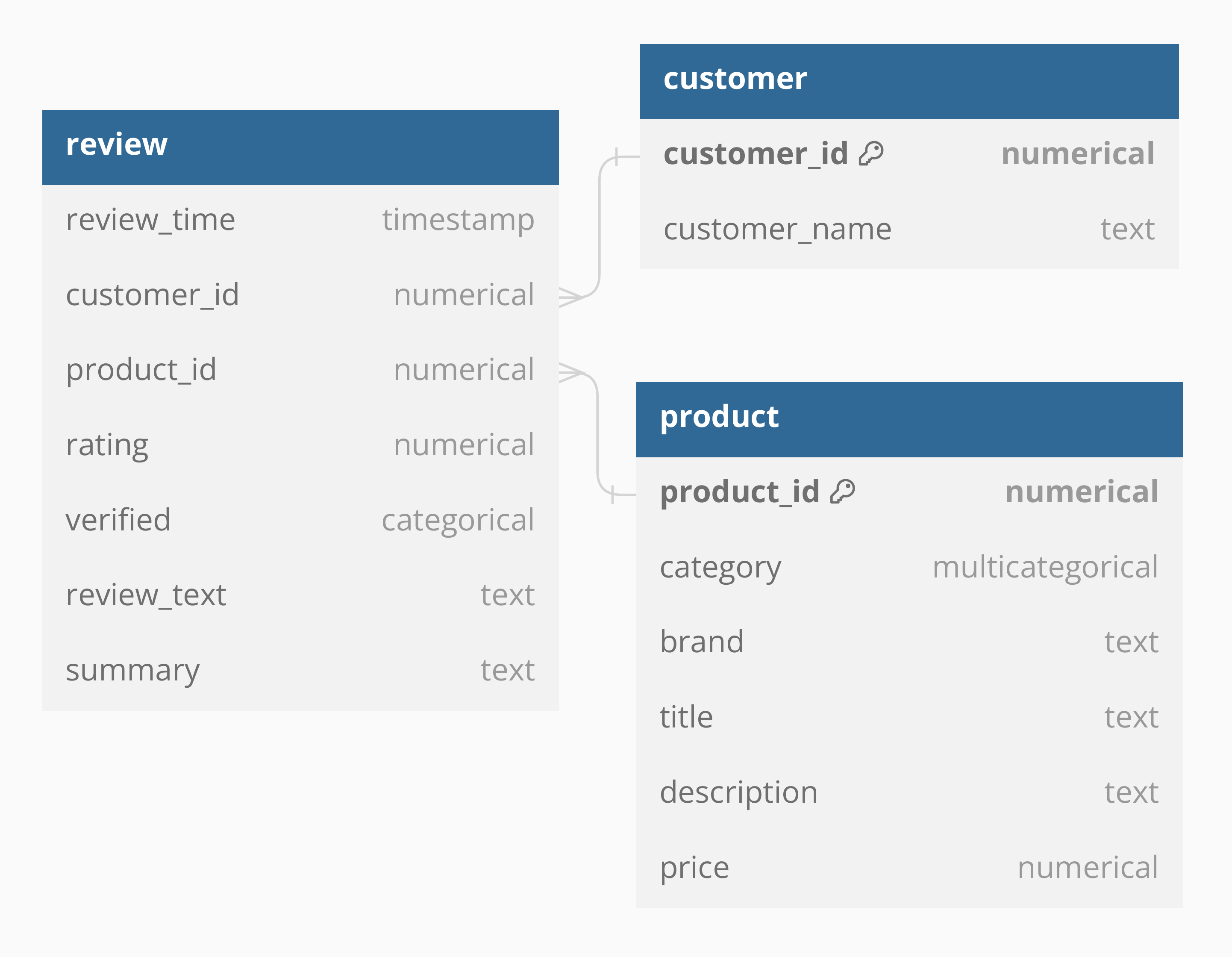}
  \caption{\Amazon contains two dimension tables (customers and products) and one fact table (reviews). Each review has a customer and a product foreign key. }
  
  \label{fig: product tables}
\end{figure}

\xhdr{Database overview} The \Amazon relational database stores product and user purchasing behavior across Amazon's e-commerce platform. Notably, it contains rich information about each product and transaction. The product table includes price and category information; the review table includes overall rating, whether the user has actually bought the product, and the text of the review itself. We use the subset of book-related products. The entity relationships are described in Fig. \ref{fig: product tables}.

\xhdr{Dataset statistics} \Amazon covers 3 relational tables and contains 1.85M customers, 21.9M reviews, 506K products. This relational database spans from 1996-06-25 to 2018-09-28. The validation timestamp $t_{\text{val}}$ is set to 2014-01-21 and the testing timestamp $t_{\text{test}}$ is 2016-01-01. Thus, tasks can have a window size up to 2 years.

\subsubsection{\AmazonLTV: Predict the life time value (LTV) of a user}

\textbf{Task definition:} Predict the life time value of a user, defined as the sum of prices of the products that the user will buy and review in the next 2 years. 

\textbf{Entity filtering:} We filter on active users defined as users that wrote review in the past two years before the timestamp. 

\textbf{Task significance:} By accurately forecasting LTV, the e-commerce platform can gain insights into user purchasing patterns and preferences, which is essential when making strategic decisions related to marketing, product recommendations, and inventory management. Understanding a user's future purchasing behavior helps in tailoring personalized shopping experiences and optimizing product assortments, ultimately enhancing customer satisfaction and loyalty.

\textbf{Machine learning task:} Regression. The target ranges from \$0-\$33,858.4 in the given time window in the training table.

\textbf{Evaluation metric:} Mean Absolute Error (MAE).

\subsubsection{\AmazonChurn: Predict if the user churns} 

\textbf{Task definition:} Predict if the user will not buy any product in the next 2 years. 

\textbf{Entity filtering:} We filter on active users defined as users that wrote review in the past two years before the timestamp. 

\textbf{Task significance:} Predicting churn accurately allows companies to identify potential risks of customer attrition early on. By understanding which customers are at risk of disengagement, businesses can implement targeted interventions to improve customer retention. This may include personalized marketing, tailored offers, or enhanced customer service. Effective churn prediction enables businesses to maintain a stable customer base, ensuring sustained revenue streams and facilitating long-term planning and resource allocation.

\textbf{Machine learning task:} Binary classification. The label is 1 when user churns and 0 vice versus. 

\textbf{Evaluation metric:} Average precision (AP).

\subsection{\StackEx: Stack exchange question-and-answer website database}\label{sec:stack}

\begin{figure}[h] % 'h' for here
  \centering
  \includegraphics[width=\linewidth]{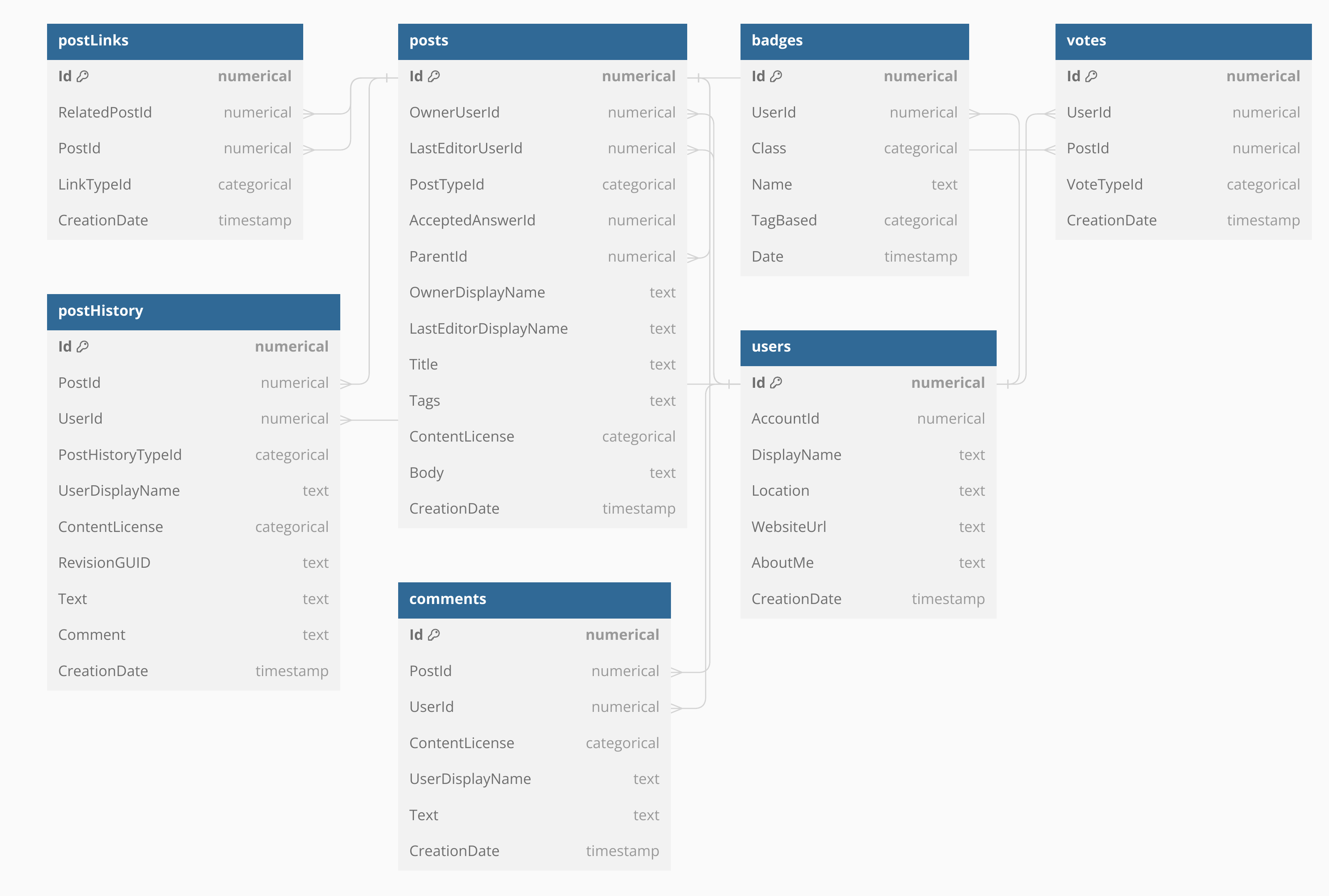}
  \caption{Entity relational diagrams of \texttt{Stack-Exchange}.}
  \label{fig:stack_exchange}
\end{figure}

\xhdr{Database overview} Stack Exchange is a network of question-and-answer websites on topics in diverse fields, each site covering a specific topic, where questions, answers, and users are subject to a reputation award process. The reputation system allows the sites to be self-moderating. In our benchmark, we use the stats-exchange site. We derive from the raw data dump from 2023-09-12. Figure~\ref{fig:stack_exchange} shows its entity relational diagrams. 

\xhdr{Dataset statistics} \StackEx covers 7 relational tables and contains 333K users, 415K posts, 794K comments, 1.67M votes, 103K post links, 590K badges records, 1.49M post history records. This relational database spans from 2009-02-02 to 2023-09-03. The validation timestamp $t_{\text{val}}$ is set to be 2019-01-01 and the testing timestamp $t_{\text{test}}$ is set to be 2021-01-01. Thus, the maximum time window size for predictive task is 2 years. 

\subsubsection{\StackExEngage: Predict if a user will be an active contributor to the site} 

\textbf{Task definition:} Predict if the user will make any contribution, defined as vote, comment, or post, to the site in the next 2 years. 

\textbf{Entity filtering:} We filter on active users defined as users that have made at least one comment/post/vote before the timestamp. 

\textbf{Task significance:} By accurately forecasting the levels of user contribution, website administrators can effectively gauge and oversee user activity. This insight allows for well-informed choices across various business aspects. For instance, it aids in preempting and mitigating user attrition, as well as in enhancing strategies to foster increased user interaction and involvement. This predictive task serves as a crucial tool in optimizing user experience and sustaining a dynamic and engaged user base.

\textbf{Machine learning task:} Binary classification. The label is 1 when user contributes to the site and 0 otherwise.

\textbf{Evaluation metric:} Average Precision (AP).

\subsubsection{\StackExVotes: Predict the number of upvotes a question will receive}

\textbf{Task definition:} Predict the popularity of a question post in the next six months. The popularity is defined as the number of upvotes the post will receive.

\textbf{Entity filtering:} We filter on question posts that are posted recently in the past 2 years before the timestamp. This ensures that we do not predict on old questions that have been outdated.

\textbf{Task significance:} Predicting the popularity of a question post is valuable as it empowers site managers to predict and prepare for the influx of traffic directed towards that particular post. This foresight is instrumental in making strategic business decisions, such as curating question recommendations and optimizing content visibility. Understanding which posts are likely to attract more attention helps in tailoring the user experience and managing resources effectively, ensuring that the most engaging and relevant content is highlighted to maintain and enhance user engagement.

\textbf{Machine learning task:} Regression. The target ranges from 0-52 number of upvotes in the given time window in the training table.

\textbf{Evaluation metric:} Mean Absolute Error (MAE).

%% file: new-challenges.tex
\section{A New Program for Graph Representation Learning}\label{sec: new challenges}

\begin{figure}[h] % 'h' for here
  \centering
  \includegraphics[width=0.9\linewidth]{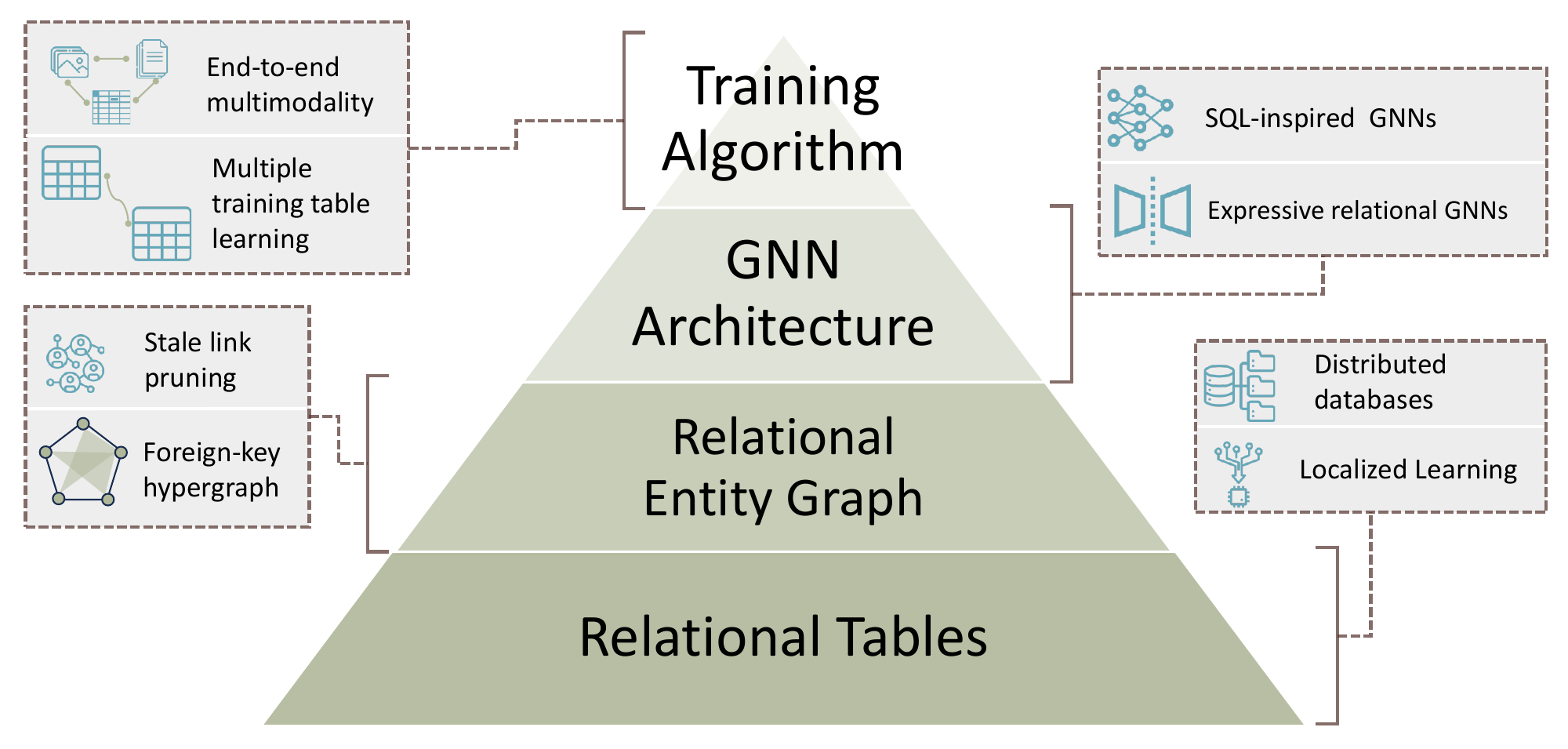}
  \caption{\AreaName brings new challenges at all levels of the machine learning stack.}
  \label{fig: research landscape}
\end{figure}

Developing \AreaName requires a new research program in graph representation learning on relational data. There are opportunities at all levels of the research stack, including (pre-)training methods, GNN architectures, multimodality, new graph formulations, and scaling to large distributed relational databases for many industrial settings. Here we discuss several promising aspects of this research program, aiming to stimulate the interest of the graph machine learning community.

\subsection{Scaling \AreaName}

Relational databases are often vast, with information distributed across many servers with constrained communication. However, relational data has a non-typical graph structure which may help scale \AreaName more efficiently.

\xhdr{Distributed Training on Relational Data} Existing distributed machine learning techniques often assume that each server contains data of the same type. Relational data on the other hand, naturally partitions in to pieces, bringing new challenges depending on the partitioning technique used. \emph{Horizontal partitioning}, known as sharding, is the most common approach. It creates database shards by splitting tables row-wise according to some criterion (e.g., all customer with zipcode in a given range). In this case, a table containing a customers personal record may lie on a distinct server from the table recording recent purchase activity, leading to communication bottlenecks when attempting to train models by sensing messages between purchases and customer records. Less common, but also possible, is vertical splitting. Different splitting options suggests an opportunity to (a) develop specialized distributed learning methods that exploit the vertical or horizontal partitioning, and (b) design further storage arrangements that may be more suited to \AreaName. The question of graph partitioning arises in all large-scale graph machine learning settings, however in this case we are fortunate to have non-typical graph structure (i.e., it follows the schema) which makes it easier to find favourable partitions.

\xhdr{Localized Learning} For many predictive tasks it is neither feasible (due to database size) nor desirable (due to task narrowness) to propagate GNN messages across the entire graph during training. In such cases, sampling schemes that preserve locality by avoiding exponential growth in GNN receptive field are needed. This is easily addressed in cases with  prior knowledge on the relevant entities. How to scale to deep models that remain biased models towards local computation in cases with no prior knowledge remains an interesting open question.

\subsection{Building Graphs from Relational Data}

An essential ingredient of \AreaName is the use of an individual entity and relation-level graph on which to apply inter-entity message passing to learn entity embeddings based on relations to other entities.  In Sec. \ref{sec:formal_rl_graph} we introduced one such graph, the \emph{relational entity graph}, a general procedure for viewing any relational database as a graph. Whilst a natural choice, we do not propose dogmatically viewing entities as nodes and relations as edges.  Instead, the essential property of the relational entity graph is that it is \emph{full-resolution}. That is, each entity and each primary-foreign key link in the relational database corresponds to its own graph-piece, so that the relational database is exactly encoded in graph form. It is this property that we expect potential alternative graph designs to share. Beyond this stipulation, many creative graph choices are possible, and we discuss some possibilities here.

\xhdr{Foreign-key Hypergraph} Fact tables often contain entities with a fixed foreign-key pattern (\emph{e.g.}, in \Amazon a row in a \tbl{review} table always refers to a \tbl{customer} and a \tbl{product} foreign key). The relational entity graph views a review as a node, with edges to a customer and product. However, another possibility is to view this as a single hyperedge between review, customer, and product. Alternative graph choices may alter (and improve) information propagation between entities (\emph{cf.} Sec. \ref{sec: new gnns}).

\xhdr{Stale Link Pruning} Entities that have been active for a long time may have a lot of links to other entities. Many of these links may be \emph{stale}, or uninformative, for certain tasks. For example, the purchasing patterns of a longtime customer during childhood are likely to be less relevant to their purchasing patterns in adulthood. Links that are stale for a certain task may hurt predictive power by obfuscating true predictive signals, and reduce model efficiency due to processing uninformative data. This situation calls for careful stale link and entity handling to focus on relevant information. Promising methods may include pruning or preaggregating stale links. More generally, how to deal with more gradual distribution drift over time is an open question.

\subsection{GNN Architectures for Relational Data}\label{sec: new gnns}

Viewing a relational database a graphs leads to graphs with structural properties that are consistent across databases. To properly exploit this structure new specialized GNN architectures are needed. Here we discuss several concrete directions for designing new architectures.

\xhdr{Expressive GNNs for Relational Data} Relational entity graphs (\emph{cf.} Sec. \ref{sec:formal_rl_graph}) obey certain structural constraints. For example, as nodes correspond to entities drawn from one of several tables, the relational entity graph is naturally $n$-partite, where $n$ is the total number of tables.  This suggests that GNNs for relational data should be designed to be capable of learning expressive decision rules over $n$-partite graphs. Unfortunately,  recent studies find that many GNN architectures fail to distinguish biconnected graphs \citep{zhang2023rethinking}. Further work is needed to design expressive $n$-partite graph models.

Relational entity graphs also have regularity in edge-connectivity. For instance, in \Amazon entities in the \tbl{review} table always refer to one \tbl{customer} and one \tbl{product}.
Consistent edge patterns are described by the structure of the schema graph $(\mathcal T, \mathcal R)$ (\emph{cf.} Sec. \ref{sec:schema_graph}). How to integrate prior knowledge of the graph structure of $(\mathcal T, \mathcal R)$ into GNNs that operate on an entity-level graph (the relational entity graph) remains an open question. These two examples serve only to illustrate the possibilities for architecture design based on the structure of relational entity graphs. Many other structural properties of relational data may lead to innovative new expressive GNN architectures.

\xhdr{Query Language Inspired Models} SQL operations are known to be extremely powerful operations for manipulating relational data. Their weakness is that they do not have differentiable parameters, making end-to-end learnability impossible. Despite this, there are close similarities between key SQL queries and the computation process of graph neural networks. For instance, a very common way to combine information across tables $T_1,T_2$ in SQL is to (1) create a table $T_3$ by applying a \texttt{JOIN} operation to table $T_1$ and $T_2$, by matching foreign keys in $T_1$ to primary keys in $T_2$, then (2) produce a final table with the same number of rows as $T_2$ by applying an \texttt{AGGREGATE} operation to rows in $T_3$ with foreign keys pointing to the same entity in $T_2$. There are many choices of \texttt{AGGREGATE} operation such as \texttt{SUM}, \texttt{MEAN}  and  \texttt{COUNT}. This process \emph{directly} mirrors GNN computations of messages from neighboring nodes, followed by message aggregation. In other words, GNNs can be thought of as a neural version of SQL \texttt{JOIN}+\texttt{AGGREGATE} operations. This suggests that an opportunity for powerful new neural network architectures by designing differentiable computation blocks that algorithmically align \citep{xu2019can} to existing SQL operations that are known to be useful.

\xhdr{New Message Passing Schemes} Beyond expressivity, new architectures may also improve information propagation between entities. For instance, collaborative filtering methods enhance predictions by identify entities with similar behavior patterns, customers with similar purchase history. However, in the relational entity graph, the two related customers may not be directly linked. Instead they are indirectly be linked to one another through links to their respective purchases, which are linked to a particular shared product ID. This means that a standard message passing GNN will require four message passing steps to propagate the information that customer $v_1$ purchased the same product as customer $v_2$ (2-hops from $v_1$ to product, and 2-hops from product to $v_1$). New message passing schemes that do multiple hops or directly connect customers (more generally, entities) with similar behavior patterns may more effectively propagate key information through the model. As well as new message passing schemes, there is also opportunity for new message aggregation methods. One possibility is order dependent aggregation, that combines messages in a time-dependent way, as explored by \cite{yang2022stam}. Another is schema dependent aggregation, that combines messages based on what part of the schema graph the messages are arriving from.

\subsection{Training Techniques for Relational Data}

By its nature, relational data contains highly overlapping predictive signals and tasks. This interconnectedness of data and tasks is a big opportunity for new neural network training methods that maximally take advantage of this interconnectedness to identify useful predictive signals. This section discusses several such opportunities.

\xhdr{Multi-Task Learning} 
 Many predictive tasks on relational data are distinct but related. For example, predicting customer lifetime value, and forecasting individual product sales both involve anticipating future purchase patterns. In {\BenchmarkName}, this corresponds to defining multiple training tables, one for each task, and training a single model jointly on all tasks in order to benefit from shared predictive signals. How to group training tables to leverage their overlap is a promising area for further study.

\xhdr{Multi-Modal Learning}

Entities in relational databases often have attributes covering multiple modalities (\emph{e.g.,} products come with images, descriptions, as well as different categorical and numerical features). The \AreaName blueprint first extracting entity-level features, which are used as initial node-features for the GNN model.
In \BenchmarkName, this multimodal entity-level feature extraction is handled by using state-of-the-art pre-trained models using the PyTorch Frame library to pre-extract features. This maximizes convenience for graph-focused research, but is likely suboptimal because the entity-level feature extraction model is frozen. This is especially relevant in contexts with unusual data---e.g., specialized medical documentation---that generic pre-trained models will likely fail to extract important details.
To facilitate exploration of joint entity-level and graph-level modelling, \BenchmarkName also provides the option to load raw data, to allow researchers to experiment with different feature-extraction methods.

\xhdr{Foundation Models and Data Type Encoders} In practice, new predictive tasks on relational data are often specified on-the-fly, with fast responses required. Such situations preclude costly model training from scratch, instead requiring powerful and generic pre-trained models. Self-supervised labels for model pre-training can be mined from historical data, just as with training table construction. However, techniques for automatically deciding which labels to mine remains unexplored. Another desirable property of pre-trained models is that they are \emph{inductive}, so they can be applied to entirely new relational databases out-of-the-box. This presents a challenges in how to deal with unseen column types and relations between tables. Such flexibility is needed in order to move towards foundation models for relational databases. More broadly, how to build column encoders is an important question. As well as distribution shifts as mentioned in the previous paragraph, there are also decisions on when to share column encoders (should two image columns use the same image encoder?), as well as special data types such as static time intervals (\emph{e.g.}, to describe the time period an employee worked at a company, or the time period in which a building project was conducted). Special data types may require specialized encoder choices, and possibly even deeper integration into the neural network computation graph. How best to aggregate of cross-modal information into a single fused embedding is another question for exploration.

%% file: related.tex
\section{Related Work}\label{sec: related work}

\AreaName touches on many areas of related work which we survey next.

\xhdr{Statistical Relational Learning} Since the foundation of the field of AI, sought to design systems capable of reasoning about entities and their relations, often by explicitly building graph structures \citep{minsky1974framework}. Each new era of AI research also brought its own form of relational learning. A prominent instance is statistical relational learning \citep{de2008logical}, a common form of which seeks to describe objects and relations in terms of first-order logic, fused with graphical models to model uncertainty \citep{getoor2001learning}. These descriptions can then be used to generate new ``knowledge'' through inductive logic programming \citep{lavrac1994inductive}. Markov logic networks, a prominent statistical relational approach, are defined by a collection of first-order logic formula with accompanying scalar weights \citep{richardson2006markov}. This information is then used to  define a probability distribution over possible worlds (via Markov random fields) which enables probabilistic reasoning about the truth of new formulae. We see \AreaName as inheriting this lineage, since both approaches operate on data with rich relational structure, and both approaches integrate relational structure into the model design. Of course, there are important distinctions between the two methods too, such as the natural scalability of graph neural network-based methods, and that \AreaName does not rely on first-order logic to describe data, allowing broad applicability to relations that are hard to fit into this form.

\xhdr{Tabular Machine Learning} Tree based methods, notably XGBoost \citep{chen2016xgboost}, remain key workhorses of enterprise machine learning systems due to their scalability and reliability. In parallel, efforts to design deep learning architectures for tabular data have continued \citep{huang2020tabtransformer,arik2021tabnet,gorishniy2021revisiting,gorishniy2022embeddings,chen2023trompt}, but have struggled to clearly establish dominance over tree-based methods \citep{shwartz2022tabular}. The vast majority of tabular machine learning focuses on the single table setting, which we argue forgoes use of the rich interconnections between relational data. As such, it does not address the key problem, which is how to get the data from a multi-table to a single table representation. Recently, a nascent body of work has begun to consider multiple tables. For instance, \cite{zhu2023xtab} pre-train tabular Transformers that generalize to new tables with unseen columns. 

\xhdr{Knowledge Graph Embedding}  Knowledge graphs store relational data, and highly scalable knowledge graph embeddings methods have been developed over the last decade to embed data into spaces whose geometry reflects the relational struture~\citep{bordes2013translating,wang2014knowledge,wang2017knowledge}. Whilst also dealing with relational data, this literature differs from this present work in the task being solved. The key task of knowledge graph embeddings is to predict missing entities (Q: Who was Yann LeCun's postdoc advisor? A: Geoffrey Hinton)  or relations  (Q: Did Geoffrey Hinton win a Turing Award? A: Yes). To assist in such \emph{completion} tasks, knowledge graph methods learn an embedding space with the goal of exactly preserving the relation semantics between entities. This is different from \AreaName, which aims to make \emph{predictions} about entities, or groups of entities. Because of this, \AreaName seeks to leverage relations to learn entity representations, but does not need to learn an embedding space that perfectly preserves all relation semantics. This gives more freedom and flexibility to our models, which may discard certain relational information it finds unhelpful. Nonetheless, adopting ideas from knowledge graph embedding may yet be fruitful.
 
\xhdr{Deep Learning on Relational Data} Proposals to use message passing neural networks on relational data have occasionally surfaced within the research community. In particular, \cite{schlichtkrull2018relational, cvitkovic2020supervisedrd, vsir2021deep}, and \cite{zahradnik2023deep} make the connection between relational data and graph neural networks and explore it with different network architectures, such as heterogeneous message passing. However, our aim is to move beyond the conceptual level, and clearly establish deep learning on relational data as a subfield of machine learning. Accordingly, we focus on the components needed to establish this new area and attract broader interest: (1) a clearly scoped design space for neural network architectures on relational data, (2) a carefully chosen suite of benchmark databases and predictive tasks around which the community can center its efforts, (3) standardized data loading and splitting, so that temporal leakage does not contaminate experimental results, (4) recognizing time as a first-class citizen, integrated into all sections of the experimental pipeline, including temporal data splitting, time-based forecasting tasks, and temporal-based message passing, and (5) standardized evaluation protocols to ensure comparability between reported results.

%% file: conclusion.tex
\section{Conclusion}

A large proportion of the worlds data is natively stored in relational tables. Fully exploiting the rich signals in relational data therefore has the potential to rewrite what problems computing can solve. We believe that \AreaName will make it possible to achieve superior performance on various prediction problems spanning the breadth of human activity, leading to considerable improvements in automated decision making.  There is currently a great scientific opportunity to develop the field of \AreaName, and further refine this vision.

This paper serves as a road map in this pursuit. We introduce a blueprint for a neural network architecture  that directly processes relational data by casting predictive tasks as graph representation learning problems. In Sec. \ref{sec: new challenges} we discuss the many new challenges and opportunities this presents for the graph machine learning community. To facilitate research, we introduce \BenchmarkName, a set of benchmark datasets, and a Python package for data loading, and model evaluation.

%% file: main.bbl
\begin{thebibliography}{48}
\providecommand{\natexlab}[1]{#1}
\providecommand{\url}[1]{\texttt{#1}}
\expandafter\ifx\csname urlstyle\endcsname\relax
  \providecommand{\doi}[1]{doi: #1}\else
  \providecommand{\doi}{doi: \begingroup \urlstyle{rm}\Url}\fi

\bibitem[Amodei et~al.(2016)Amodei, Ananthanarayanan, Anubhai, Bai, Battenberg,
  Case, Casper, Catanzaro, Cheng, Chen, et~al.]{amodei2016deep}
Dario Amodei, Sundaram Ananthanarayanan, Rishita Anubhai, Jingliang Bai, Eric
  Battenberg, Carl Case, Jared Casper, Bryan Catanzaro, Qiang Cheng, Guoliang
  Chen, et~al.
\newblock Deep speech 2: End-to-end speech recognition in english and mandarin.
\newblock In \emph{International Conference on Machine Learning (ICML)}, 2016.

\bibitem[Arik and Pfister(2021)]{arik2021tabnet}
Sercan~{\"O} Arik and Tomas Pfister.
\newblock Tabnet: Attentive interpretable tabular learning.
\newblock In \emph{Proceedings of the AAAI conference on artificial
  intelligence}, volume~35, pages 6679--6687, 2021.

\bibitem[Bordes et~al.(2013)Bordes, Usunier, Garcia-Duran, Weston, and
  Yakhnenko]{bordes2013translating}
Antoine Bordes, Nicolas Usunier, Alberto Garcia-Duran, Jason Weston, and Oksana
  Yakhnenko.
\newblock Translating embeddings for modeling multi-relational data.
\newblock In \emph{Advances in Neural Information Processing Systems
  (NeurIPS)}, 2013.

\bibitem[Brown et~al.(2020)Brown, Mann, Ryder, Subbiah, Kaplan, Dhariwal,
  Neelakantan, Shyam, Sastry, Askell, et~al.]{brown2020language}
Tom~B Brown, Benjamin Mann, Nick Ryder, Melanie Subbiah, Jared Kaplan, Prafulla
  Dhariwal, Arvind Neelakantan, Pranav Shyam, Girish Sastry, Amanda Askell,
  et~al.
\newblock Language models are few-shot learners.
\newblock In \emph{Advances in Neural Information Processing Systems
  (NeurIPS)}, 2020.

\bibitem[Chamberlin and Boyce(1974)]{chamberlin1974sequel}
Donald~D Chamberlin and Raymond~F Boyce.
\newblock Sequel: A structured english query language.
\newblock In \emph{Proceedings of the 1974 ACM SIGFIDET (now SIGMOD) workshop
  on Data description, access and control}, pages 249--264, 1974.

\bibitem[Chen et~al.(2023)Chen, Chiang, Chou, Chen, and Chang]{chen2023trompt}
Kuan-Yu Chen, Ping-Han Chiang, Hsin-Rung Chou, Ting-Wei Chen, and Tien-Hao
  Chang.
\newblock Trompt: Towards a better deep neural network for tabular data.
\newblock In \emph{International Conference on Machine Learning (ICML)}, 2023.

\bibitem[Chen and Guestrin(2016)]{chen2016xgboost}
Tianqi Chen and Carlos Guestrin.
\newblock Xgboost: A scalable tree boosting system.
\newblock In \emph{ACM SIGKDD Conference on Knowledge Discovery and Data Mining
  (KDD)}, pages 785--794, 2016.

\bibitem[Codd(1970)]{codd1970relational}
Edgar~F Codd.
\newblock A relational model of data for large shared data banks.
\newblock \emph{Communications of the ACM}, 13\penalty0 (6):\penalty0 377--387,
  1970.

\bibitem[Cvitkovic(2019)]{cvitkovic2020supervisedrd}
Milan Cvitkovic.
\newblock Supervised learning on relational databases with graph neural
  networks.
\newblock \emph{ICLR Workshop on Representation Learning on Graphs and
  Manifolds}, 2019.

\bibitem[{DB-Engines}(2023)]{db-engines}
{DB-Engines}.
\newblock {DBMS popularity broken down by database model}, 2023.
\newblock Available: \url{https://db-engines.com/en/ranking_categories}.

\bibitem[De~Raedt(2008)]{de2008logical}
Luc De~Raedt.
\newblock \emph{Logical and relational learning}.
\newblock Springer Science \& Business Media, 2008.

\bibitem[Devlin et~al.(2018)Devlin, Chang, Lee, and Toutanova]{devlin2018bert}
Jacob Devlin, Ming-Wei Chang, Kenton Lee, and Kristina Toutanova.
\newblock Bert: Pre-training of deep bidirectional transformers for language
  understanding.
\newblock In \emph{North American Chapter of the Association for Computational
  Linguistics (NAACL)}, 2018.

\bibitem[Fey and Lenssen(2019)]{fey2019fast}
Matthias Fey and Jan~Eric Lenssen.
\newblock Fast graph representation learning with pytorch geometric.
\newblock \emph{ICLR 2019 (RLGM Workshop)}, 2019.

\bibitem[Garcia-Molina et~al.(2008)Garcia-Molina, Ullman, and
  Widom]{ullman-book}
Hector Garcia-Molina, Jeffrey~D. Ullman, and Jennifer Widom.
\newblock \emph{Database Systems: The Complete Book}.
\newblock Prentice Hall Press, USA, 2 edition, 2008.
\newblock ISBN 9780131873254.

\bibitem[Geirhos et~al.(2020)Geirhos, Jacobsen, Michaelis, Zemel, Brendel,
  Bethge, and Wichmann]{geirhos2020shortcut}
Robert Geirhos, J{\"o}rn-Henrik Jacobsen, Claudio Michaelis, Richard Zemel,
  Wieland Brendel, Matthias Bethge, and Felix~A Wichmann.
\newblock Shortcut learning in deep neural networks.
\newblock \emph{Nature Machine Intelligence}, 2\penalty0 (11):\penalty0
  665--673, 2020.

\bibitem[Getoor et~al.(2001)Getoor, Friedman, Koller, and
  Pfeffer]{getoor2001learning}
Lise Getoor, Nir Friedman, Daphne Koller, and Avi Pfeffer.
\newblock Learning probabilistic relational models.
\newblock \emph{Relational data mining}, pages 307--335, 2001.

\bibitem[Gilmer et~al.(2017)Gilmer, Schoenholz, Riley, Vinyals, and
  Dahl]{gilmer2017mpgnn}
Justin Gilmer, Samuel~S. Schoenholz, Patrick~F. Riley, Oriol Vinyals, and
  George~E. Dahl.
\newblock Neural message passing for quantum chemistry.
\newblock In \emph{International Conference on Machine Learning (ICML)}, page
  1263–1272, 2017.

\bibitem[Gorishniy et~al.(2021)Gorishniy, Rubachev, Khrulkov, and
  Babenko]{gorishniy2021revisiting}
Yury Gorishniy, Ivan Rubachev, Valentin Khrulkov, and Artem Babenko.
\newblock Revisiting deep learning models for tabular data.
\newblock In \emph{Advances in Neural Information Processing Systems
  (NeurIPS)}, volume~34, pages 18932--18943, 2021.

\bibitem[Gorishniy et~al.(2022)Gorishniy, Rubachev, and
  Babenko]{gorishniy2022embeddings}
Yury Gorishniy, Ivan Rubachev, and Artem Babenko.
\newblock On embeddings for numerical features in tabular deep learning.
\newblock \emph{Advances in Neural Information Processing Systems},
  35:\penalty0 24991--25004, 2022.

\bibitem[Hamilton et~al.(2017)Hamilton, Ying, and
  Leskovec]{hamilton2017inductive}
Will Hamilton, Zhitao Ying, and Jure Leskovec.
\newblock Inductive representation learning on large graphs.
\newblock In \emph{Advances in Neural Information Processing Systems
  (NeurIPS)}, 2017.

\bibitem[Hannun et~al.(2014)Hannun, Case, Casper, Catanzaro, Diamos, Elsen,
  Prenger, Satheesh, Sengupta, Coates, et~al.]{hannun2014deep}
Awni Hannun, Carl Case, Jared Casper, Bryan Catanzaro, Greg Diamos, Erich
  Elsen, Ryan Prenger, Sanjeev Satheesh, Shubho Sengupta, Adam Coates, et~al.
\newblock Deep speech: Scaling up end-to-end speech recognition.
\newblock \emph{arXiv preprint arXiv:1412.5567}, 2014.

\bibitem[He et~al.(2016)He, Zhang, Ren, and Sun]{he2016deep}
Kaiming He, Xiangyu Zhang, Shaoqing Ren, and Jian Sun.
\newblock Deep residual learning for image recognition.
\newblock In \emph{IEEE Conference on Computer Vision and Pattern Recognition
  (CVPR)}, pages 770--778, 2016.

\bibitem[Hu et~al.(2023)Hu, Fey, Yuan, Zhang, Nitta, Cao, and
  Kocijan]{Hu_PyTorch_Frame_A_2023}
Weihua Hu, Matthias Fey, Yiwen Yuan, Zecheng Zhang, Akihiro Nitta, Kaidi Cao,
  and Vid Kocijan.
\newblock {PyTorch Frame: A Deep Learning Framework for Tabular Data}, October
  2023.
\newblock URL \url{https://github.com/pyg-team/pytorch-frame}.

\bibitem[Hu et~al.(2020)Hu, Dong, Wang, and Sun]{hu2020hgt}
Ziniu Hu, Yuxiao Dong, Kuansan Wang, and Yizhou Sun.
\newblock Heterogeneous graph transformer.
\newblock In \emph{Proceedings of The Web Conference 2020}, page 2704–2710,
  2020.

\bibitem[Huang et~al.(2020)Huang, Khetan, Cvitkovic, and
  Karnin]{huang2020tabtransformer}
Xin Huang, Ashish Khetan, Milan Cvitkovic, and Zohar Karnin.
\newblock Tabtransformer: Tabular data modeling using contextual embeddings.
\newblock \emph{arXiv preprint arXiv:2012.06678}, 2020.

\bibitem[Johnson et~al.(2016)Johnson, Pollard, Shen, Lehman, Feng, Ghassemi,
  Moody, Szolovits, Anthony~Celi, and Mark]{johnson2016mimic}
Alistair~EW Johnson, Tom~J Pollard, Lu~Shen, Li-wei~H Lehman, Mengling Feng,
  Mohammad Ghassemi, Benjamin Moody, Peter Szolovits, Leo Anthony~Celi, and
  Roger~G Mark.
\newblock Mimic-iii, a freely accessible critical care database.
\newblock \emph{Scientific data}, 3\penalty0 (1):\penalty0 1--9, 2016.

\bibitem[{Kaggle}(2022)]{kaggle-survey}
{Kaggle}.
\newblock {Kaggle Data Science \& Machine Learning Survey}, 2022.
\newblock Available:
  \url{https://www.kaggle.com/code/paultimothymooney/kaggle-survey-2022-all-results/notebook}.

\bibitem[Kapoor and Narayanan(2023)]{kapoor2023leakage}
Sayash Kapoor and Arvind Narayanan.
\newblock Leakage and the reproducibility crisis in machine-learning-based
  science.
\newblock \emph{Patterns}, 4\penalty0 (9), 2023.

\bibitem[Lavrac and Dzeroski(1994)]{lavrac1994inductive}
Nada Lavrac and Saso Dzeroski.
\newblock Inductive logic programming.
\newblock In \emph{WLP}, pages 146--160. Springer, 1994.

\bibitem[Minsky(1974)]{minsky1974framework}
Marvin Minsky.
\newblock A framework for representing knowledge, 1974.

\bibitem[{PubMed}(1996)]{pubmed}
{PubMed}.
\newblock {National Center for Biotechnology Information, U.S. National Library
  of Medicine}, 1996.
\newblock Available: \url{https://www.ncbi.nlm.nih.gov/pubmed/}.

\bibitem[Reimers and Gurevych(2019)]{reimers2019sentence}
Nils Reimers and Iryna Gurevych.
\newblock Sentence-bert: Sentence embeddings using siamese bert-networks.
\newblock \emph{arXiv preprint arXiv:1908.10084}, 2019.

\bibitem[Richardson and Domingos(2006)]{richardson2006markov}
Matthew Richardson and Pedro Domingos.
\newblock Markov logic networks.
\newblock \emph{Machine learning}, 62:\penalty0 107--136, 2006.

\bibitem[Rossi et~al.(2020)Rossi, Chamberlain, Frasca, Eynard, Monti, and
  Bronstein]{rossi2020temporal}
Emanuele Rossi, Ben Chamberlain, Fabrizio Frasca, Davide Eynard, Federico
  Monti, and Michael Bronstein.
\newblock Temporal graph networks for deep learning on dynamic graphs.
\newblock \emph{ICML Workshop on Graph Representation Learning 2020}, 2020.

\bibitem[Russakovsky et~al.(2015)Russakovsky, Deng, Su, Krause, Satheesh, Ma,
  Huang, Karpathy, Khosla, Bernstein, et~al.]{russakovsky2015imagenet}
Olga Russakovsky, Jia Deng, Hao Su, Jonathan Krause, Sanjeev Satheesh, Sean Ma,
  Zhiheng Huang, Andrej Karpathy, Aditya Khosla, Michael Bernstein, et~al.
\newblock Imagenet large scale visual recognition challenge.
\newblock \emph{International journal of computer vision}, 115\penalty0
  (3):\penalty0 211--252, 2015.

\bibitem[Schlichtkrull et~al.(2018)Schlichtkrull, Kipf, Bloem, van den Berg,
  Titov, and Welling]{schlichtkrull2018relational}
Michael Schlichtkrull, Thomas~N. Kipf, Peter Bloem, Rianne van den Berg, Ivan
  Titov, and Max Welling.
\newblock Modeling relational data with graph convolutional networks.
\newblock In Aldo Gangemi, Roberto Navigli, Maria-Esther Vidal, Pascal Hitzler,
  Rapha{\"e}l Troncy, Laura Hollink, Anna Tordai, and Mehwish Alam, editors,
  \emph{The Semantic Web}, pages 593--607, Cham, 2018. Springer International
  Publishing.

\bibitem[Shwartz-Ziv and Armon(2022)]{shwartz2022tabular}
Ravid Shwartz-Ziv and Amitai Armon.
\newblock Tabular data: Deep learning is not all you need.
\newblock \emph{Information Fusion}, 81:\penalty0 84--90, 2022.

\bibitem[{\v{S}}{\'\i}r(2021)]{vsir2021deep}
Gustav {\v{S}}{\'\i}r.
\newblock \emph{Deep Learning with Relational Logic Representations}.
\newblock Czech Technical University, 2021.

\bibitem[Varma and Zisserman(2005)]{varma2005statistical}
Manik Varma and Andrew Zisserman.
\newblock A statistical approach to texture classification from single images.
\newblock \emph{International journal of computer vision}, 62:\penalty0 61--81,
  2005.

\bibitem[Vaswani et~al.(2017)Vaswani, Shazeer, Parmar, Uszkoreit, Jones, Gomez,
  Kaiser, and Polosukhin]{vaswani2017attention}
Ashish Vaswani, Noam Shazeer, Niki Parmar, Jakob Uszkoreit, Llion Jones,
  Aidan~N Gomez, Lukasz Kaiser, and Illia Polosukhin.
\newblock Attention is all you need.
\newblock In \emph{Advances in Neural Information Processing Systems
  (NeurIPS)}, 2017.

\bibitem[Wang et~al.(2017)Wang, Mao, Wang, and Guo]{wang2017knowledge}
Quan Wang, Zhendong Mao, Bin Wang, and Li~Guo.
\newblock Knowledge graph embedding: A survey of approaches and applications.
\newblock \emph{IEEE Transactions on Knowledge and Data Engineering},
  29\penalty0 (12):\penalty0 2724--2743, 2017.

\bibitem[Wang et~al.(2021)Wang, Cai, Liang, Ding, Wang, and
  Hooi]{wang2021temporalsampling}
Yiwei Wang, Yujun Cai, Yuxuan Liang, Henghui Ding, Changhu Wang, and Bryan
  Hooi.
\newblock Time-aware neighbor sampling for temporal graph networks.
\newblock In \emph{arXiv pre-print}, 2021.

\bibitem[Wang et~al.(2014)Wang, Zhang, Feng, and Chen]{wang2014knowledge}
Zhen Wang, Jianwen Zhang, Jianlin Feng, and Zheng Chen.
\newblock Knowledge graph embedding by translating on hyperplanes.
\newblock In \emph{Proceedings of the AAAI conference on artificial
  intelligence}, volume~28, 2014.

\bibitem[Xu et~al.(2020)Xu, Li, Zhang, Du, Kawarabayashi, and
  Jegelka]{xu2019can}
Keyulu Xu, Jingling Li, Mozhi Zhang, Simon~S Du, Ken-ichi Kawarabayashi, and
  Stefanie Jegelka.
\newblock What can neural networks reason about?
\newblock In \emph{International Conference on Learning Representations
  (ICLR)}, 2020.

\bibitem[Yang et~al.(2022)Yang, Ding, Xu, Yang, and Tang]{yang2022stam}
Zhen Yang, Ming Ding, Bin Xu, Hongxia Yang, and Jie Tang.
\newblock Stam: A spatiotemporal aggregation method for graph neural
  network-based recommendation.
\newblock In \emph{Proceedings of the ACM Web Conference 2022}, pages
  3217--3228, 2022.

\bibitem[Zahradn{\'\i}k et~al.(2023)Zahradn{\'\i}k, Neumann, and
  {\v{S}}{\'\i}r]{zahradnik2023deep}
Luk{\'a}{\v{s}} Zahradn{\'\i}k, Jan Neumann, and Gustav {\v{S}}{\'\i}r.
\newblock A deep learning blueprint for relational databases.
\newblock In \emph{NeurIPS 2023 Second Table Representation Learning Workshop},
  2023.

\bibitem[Zhang et~al.(2023)Zhang, Luo, Wang, and He]{zhang2023rethinking}
Bohang Zhang, Shengjie Luo, Liwei Wang, and Di~He.
\newblock Rethinking the expressive power of gnns via graph biconnectivity.
\newblock In \emph{International Conference on Learning Representations
  (ICLR)}, 2023.

\bibitem[Zhu et~al.(2023)Zhu, Shi, Erickson, Li, Karypis, and
  Shoaran]{zhu2023xtab}
Bingzhao Zhu, Xingjian Shi, Nick Erickson, Mu~Li, George Karypis, and Mahsa
  Shoaran.
\newblock Xtab: Cross-table pretraining for tabular transformers.
\newblock In \emph{International Conference on Machine Learning (ICML)}, 2023.

\end{thebibliography}
